\documentclass[10pt,journal,compsoc]{IEEEtran}
\ifCLASSOPTIONcompsoc
  \usepackage[nocompress]{cite}
\else
  \usepackage{cite}
\fi
\ifCLASSINFOpdf
\else
\fi
\usepackage{booktabs} % For formal tables
\usepackage{multirow}
\usepackage{graphicx}
\usepackage{enumerate}
\usepackage{subfigure}
\usepackage{threeparttable}
\usepackage{enumerate}
\usepackage{amssymb}
\usepackage[ruled,linesnumbered]{algorithm2e}
\usepackage{amsmath}
\usepackage{float}
\usepackage{setspace}
\usepackage{enumitem}
\usepackage{color,xcolor}
\usepackage{verbatim}
\usepackage{balance}
\usepackage{soul}
\usepackage{hyperref}
\usepackage{amsmath}

\newtheorem{proof}{Proof}%[section]
\begin{document}
\title{Predicting Human Mobility via Self-supervised Disentanglement Learning}%Self-supervised 
%\title{Graph-based Human Mobility Prediction via Disentanglement Learning}

\author{
        Qiang~Gao,
        Jinyu~Hong,
        Xovee~Xu,
        Ping~Kuang,
        Fan~Zhou,
        and Goce~Trajcevski
\thanks{Fan~Zhou is the corresponding author.}
%\thanks{This work was supported in part by the National Natural Science Foundation of China under Grant 62102326 and in part by the Fundamental Research Funds for the Central Universities under Grant XXX.}
\IEEEcompsocitemizethanks{
\IEEEcompsocthanksitem Qiang Gao is with the School of Computing and Artificial Intelligence, Southwestern University of Finance and Economics, Chengdu, China.\protect\\ E-mail: qianggao@swufe.edu.cn
\IEEEcompsocthanksitem Jinyu Hong, Xovee Xu, Ping Kuang, and Fan Zhou are with the University of Electronic Science and Technology, Chengdu, China.
E-mail: \{jinyuhong@std., xovee@std., kuangping@, fan.zhou@\}uestc.edu.cn
\IEEEcompsocthanksitem Goce~Trajcevski is with the Iowa State University, Iowa, USA. E-mail: gocet25@iastate.edu
}
\thanks{Manuscript received XX XX, 2022; revised XX XX, 2022.}\\
}


\markboth{IEEE Transactions on XXX,~Vol.~XX, No.~X, November~2022}%
{Shell \MakeLowercase{\textit{Gao et al.}}: Predicting Human Mobility via Self-supervised Disentanglement Learning}



\IEEEtitleabstractindextext{%
\begin{abstract}
Deep neural networks have recently achieved considerable improvements in learning human behavioral patterns and individual preferences from massive spatial-temporal trajectories data. However, most of the existing research concentrates on fusing different semantics underlying sequential trajectories for mobility pattern learning which, in turn, yields a narrow perspective on comprehending human intrinsic motions. In addition, the inherent sparsity and under-explored heterogeneous collaborative items pertaining to human check-ins hinder the potential exploitation of human diverse periodic regularities as well as common interests. Motivated by recent advances in disentanglement learning, in this study we propose a novel disentangled solution called SSDL for tackling the next POI prediction problem. SSDL primarily seeks to disentangle the potential time-invariant and time-varying factors into different latent spaces from massive trajectories data, providing an interpretable view to understand the intricate semantics underlying human diverse mobility representations. To address the data sparsity issue, we present two realistic trajectory augmentation approaches to enhance the understanding of both the human intrinsic periodicity and constantly-changing intents. In addition, we devise a POI-centric graph structure to explore heterogeneous collaborative signals underlying historical check-ins. Extensive experiments conducted on four real-world datasets demonstrate that our proposed SSDL significantly outperforms the state-of-the-art approaches -- for example, it yields up to 8.57\% improvements on ACC@1.

\end{abstract}

}

\maketitle

\IEEEdisplaynontitleabstractindextext

% \fi
\IEEEpeerreviewmaketitle

\section{Introduction}
\label{Introduction}

The proliferation of geo-tagged social media (GTSM) such as Foursquare and WeChat have enabled numerous users to post interesting places, report daily activities, and make like-minded friends, resulting in the accumulation of massive amounts of contextual data (e.g., check-ins). This, %%It, 
in turn, offers unprecedented opportunities to explore human diverse life experiences (e.g., mobility patterns) and facilitate the development of various user-centric downstream applications such as %%, e.g., 
trajectory identification~\cite{feng2022}, POI recommendation/prediction~\cite{xue2021mobtcast}, itinerary prediction~\cite{luo2022rlmob} -- to name a few.
As a fundamental task in mining check-in data, predicting \textit{human mobility} (often exemplified as %%a.k.a. 
next POI prediction/recommendation) is critical for researchers and practitioners to explore the informative semantics and mutual interactions behind human check-ins~\cite{zhao2020go,wang2020deep}. For instance, it enables one to precisely ascertain users' future intentions and draw in more potential customers for new ventures~\cite{gao2019predicting,zang2021cha}.

%%As 
Spatio-temporal check-in sequences (i.e., trajectories) reflect human daily activities upon a set of POIs, which may %%as well as 
include certain (periodic) regularities. %%, t
The majority of the pioneering works in human mobility prediction aimed at modeling human sequential behaviors %%while 
taking into account spatio-temporal preferences. For instance, in order to predict where a certain user will go in the near future, conventional approaches such as Markov Chain~\cite{mathew2012predicting} and Tensor-based Factorization~\cite{massimo2018harnessing} that rely on data-driven paradigms, attempt to incorporate individual visiting preferences and explore\textit{sequential} patterns.
However, %%Unfortunately, 
these approaches depend %%ing 
%%on 
heavily on hand-crafted characteristics and face the %%are 
challenge %%ing to 
of comprehending the diverse semantics underlying massive volumes of human trajectories. This, in turn, %%which 
leads to narrow solutions in disclosing human implicit interactive hints/signals regarding historical check-ins.

%%With the 
More recent deep learning techniques such as recurrent neural networks (RNNs) have brought about encouraging achievements of learning informative check-ins (including POIs) from human trajectories and %%, recent deep learning techniques such as recurrent neural networks (RNNs) 
become a widespread and popular %%dominant 
methodology %%solution 
in tackling miscellaneous mobility learning tasks~\cite{zhao2020go,wu2020personalized,rao2022graph}. For example, Wu et al.~\cite{wu2020personalized} present a PLSPL model, which leverages a Long-Short Term Memory (LSTM) neural network to model human short-term sequential preferences while learning contextual features of POIs behind human historical check-ins via attention mechanism. To consider the spatial and temporal influences for next POI recommendation, Kong et al.~\cite{kong2018hst} incorporate the spatial and temporal intervals between two successive check-ins into recurrent hidden states to mitigate the data sparsity of human trajectories. 
Due to the higher model efficiency and the ability to quantify the contribution of each check-in in a given trajectory, several attention mechanisms like, for example, %% such as 
self-attention and vanilla attention %%are 
emerged for %%to 
handling long human historical trajectories~\cite{gao2019predicting,luo2021stan,xue2021mobtcast}. 

%%Furthermore, s
Several state-of-the-art methods have %%strive to 
employed graph structure learning to explicitly uncover spatial correlations or collaborative signals to understand individual human interests. More concretely, they attempt to acquire expressive POI representations by considering the rich contexts of highly correlated POIs. For instance, conventional methods such as word2vec-based~\cite{feng2017poi2vec,zhao2017geo} and deepwalk-based~\cite{huang2020dan,gao2022contextual} have successfully uncovered the higher-order correlations between consecutive check-ins and offered contextual POI representations. Other schemes using popular graph neural networks (GNNs), such as graph convolutional networks~\cite{rao2022graph} and graph attention networks~\cite{lim2020stp}, primarily seek to incorporate the POI-to-POI correlations (e.g., geographical proximity) behind massive human trajectories. %%, such as geographical closeness.

Despite the recent achievements in deep human mobility learning, we observe %%argue 
that existing solutions still have three significant drawbacks:

\textit{(a) Implicit semantic entanglement.} Although there is a large body of work on human mobility representation learning, the most common scheme is to take a past check-in sequence as input, and either that sequence or the user's next POI is used as the supervision signal. The former can be framed as self-supervised learning, while the latter is standard supervised learning. Nevertheless, %%they 
both ultimately focus on fusing multiple semantics behind sequential trajectories to predict the user's next POI, which could lead to a myopic perspective and produce a non-diverse recommendation result. We call this phenomenon \textit{semantic entanglement}. In practice, human trajectories, as typical sequential data, contain rich user mobility patterns that reflect diverse periodic regularities or behavioral habits of humans. More importantly, the intrinsic individual patterns/habits of humans are difficult to change over time, but their near-term intentions/behaviors are prone to be influenced/dictated by certain time instants. %%dictates. 
Thus, we consider that human mobility patterns can be implicitly disentangled into two aspects: %%, i.e., 
\textit{time-independent} and \textit{time-dependent} behaviors. Existing solutions rely on data-driven models to understand limited mobility patterns, which fail to reveal the nature of human visiting intents. As a result, they only provide %%us with 
a narrow scope to become familiar with human future behaviors, which usually carries the risk of prediction bias due to the limited scale of trajectory data available. %%we have.

\textit{(b) Sparsity in representation learning.} Only when a user decides or is willing to check in via location-based applications can a POI be recorded, which inevitably leads to the sparsity problem when gathering historical human footprints. As a result, the sparsity problem hinders the model from learning a good representation of human mobility. 
% drastically raises the stakes for obtaining good representations of human mobility. 
% The majority of the currently employed 
Existing methods either use the next POI as the sole supervisor or complement the representation learning in a semi-supervised manner with unlabeled trajectories. These paradigms mostly follow the merit of text representation in the field of natural language processing (NLP) %%field, 
which, however, easily fails in capturing the innate rules underlying human trajectories such as individual periodic regularity.

\textit{(c) Heterogeneous collaborative signals.} Most existing efforts concentrate on learning POI-to-POI relationships (a.k.a. the connectivity of POIs) from a large number of trajectories, such as consecutive correlation and geographic proximity~\cite{lim2020stp,ijcai2021p206,gao2022contextual}. Despite the successful collaboration of individual human interests via these homogeneous graph structure learning, a notable limitation is that heterogeneous semantics affiliated with the POIs are not investigated well, yielding a limit in the exploration of affluent common preferences behind human diverse trajectories. For example, people may have similar visit time preferences for certain POIs, such as going to a café after lunch. In addition, each POI is associated with a textual description (e.g., POI category), reflecting the underlying human activity interest. We conjecture that incorporating the heterogeneous correlations between POI and their category can provide us with a coarse-grained view of the higher-order connective between POIs. For example, people often go to several fashion stores to buy clothes at a time.

%\textcolor{red}{motivations}
To address the aforementioned limitations, we present a novel solution called \textbf{SSDL}, a self-supervised disentanglement learning framework for understanding human mobility. Rather than previous data-driven representation learning, SSDL performing self-supervision in the latent space aims at seeking a clean separation of the time-independent and time-dependent vectors for diverse human trajectories, which is inspired by the recent advances of variational inference and contrastive learning. Specifically, SSDL operates the sequential variational autoencoder (VAE) with a mutual information regularization to guide the training of evidence lower bound (ELBO), aiming at promoting the disentanglement of human mobility-related representations. In particular, we provide two realistic trajectory augmentation strategies to alleviate the sparsity issue in representation learning, which can further help us enhance the understanding of human intrinsic periodicity and constantly-changing intents. In addition, we also present a POI-centric graph structure to explore human common interests underlying diverse check-ins, which primarily seeks human consecutive, geospatial, temporal-aspect, and activity-aspect interests. In sum, our contributions can be summarized as follows:
\begin{itemize}
    \item We introduce a novel disentangled representation learning framework to understand human time-independent and time-dependent behaviors of their individual mobility patterns. To the best of our knowledge, this study is the first work to disentangle human mobility and investigate how it can be used for the prediction of the next POI.
    \item We propose two practical trajectory augmentation methods, guided by the inherent characteristics of individual human mobility patterns, to promote disentanglement learning.
    \item To capture heterogeneous collaborative signals behind historical check-ins, We devise a flexible POI-centric network structure to explore rich human interests in trajectories, which enhances the performance of downstream next POI prediction task.
    \item We conduct extensive experiments on four real-world datasets to evaluate the performance of our proposed SSDL. The results demonstrate that our approach outperforms state-of-the-art methods.
\end{itemize}

\section{Related Work}
\label{Related_Work}

\subsection{Next POI Prediction in Deep Learning}

Recent deep learning solutions have stimulated many researchers and practitioners to learn human periodic regularities from massive historical check-ins. Especially, deep (recurrent) neural networks such as LSTM~\cite{hochreiter1997long} and GRU~\cite{cho2014learning} have received widespread interest in the next POI prediction task as they are able to capture the sequential dependencies for mobility pattern understanding. For instance, \cite{liu2016predicting} extends the vanilla RNN model and integrates the spatial-temporal impacts into each RNN cell, yielding promising results on the next location prediction. 
Zhao et al. propose a novel ST-LSTM that implements time gates and distance gates into standard LSTM, aiming at capturing the spatio-temporal relation between consecutive check-ins~\cite{zhao2020go}. To learn more contextual information, Wu et al.~\cite{wu2020personalized} propose a personalized long- and short-term preference learning scheme to learn the specific user context, where the different influences of locations and categories of POIs are considered. While most of endeavors focus on pruning or modifying the RNN-based modules~\cite{yu2020category,zhao2020discovering,sun2021mfnp}, researches also tried to adopt other popular deep neural networks for next POI prediction, e.g., attention-based neural networks~\cite{zhang2021snpr,xue2021mobtcast} and convolutional neural networks~\cite{gao2019predicting,miao2020predicting}. Xue et al.~\cite{xue2021mobtcast} build the Transformer architecture as the mobility feature extractor in which it regards the historical trajectory and semantic contexts as the input to handle multiple factors such as temporal and geographic contexts. 
\subsection{Mobility Representation Learning}

POI embedding and trajectory embedding, as two core components in mobility representation learning, have been investigated in recent studies.

For POI embeddings, the earlier studies such as \cite{liu2016predicting} and \cite{chen2020context} set a fixed or learnable matrix as the initial representations of POIs, primarily seeking to alleviate the ``Curse of Dimensionality'' concern. However, any semantic information between POIs is under-explored. As word embeddings, especially word2vec-based~\cite{mikolov2013efficient}, have achieved great performance in NLP, recent studies also proposed various word2vec-based solutions aimed at capturing the proximity semantics of POIs from human check-in sequences (or real-world trajectories). For instance, \cite{liu2016exploring} and \cite{gao2017identifying} regard each POI as a ``word'' while each human trajectory as a ``sequence'', and use word2vec to obtain a low-dimensional vector for each POI. POI2Vec is a latent representation model that incorporates geographic influence when using word2vec method for POI embedding~\cite{feng2017poi2vec}. However, training sparse trajectories to obtain POI representations often confronts the problem of poor capability of POI semantics. 
More recently, the extraordinary success of graph neural networks (GNNs) has inspired tremendous researchers to turn to devise graph-based models to facilitate the learning of human trajectories~\cite{rao2022graph,lim2020stp}. For instance, \cite{lim2020stp} proposes a graph-based model to explore the spatial, temporal, and preference factors behind the POIs. However, it only considers homogeneous interactions among the POIs and ignores heterogeneous interactions with other key entities such as activity and check-in time.

Regarding trajectory representation learning, the majority of existing research concentrates on taking the historical trajectory as input and using the next POI as the sole supervision signal~\cite{zhao2020go,zang2021cha,liu2016predicting,yang2022getnext}. To address the narrow scale of trajectory data, some efforts attempt to employ the unlabeled trajectories as supplements and train them with the labeled trajectories jointly in an unsupervised or self-supervised manner to acquire a good representation for each trajectory~\cite{gao2019predicting,zhou2021self,zhou2021metamove}. 
Especially, to operate the trajectories in a latent space, recent studies employ generative models such as variational inference or adversarial models to learn the intrinsic distribution underlying massive trajectory data and then turn to fine-tune the model for the next POI prediction tasks. For instance, VANext extended the variational autoencoder (VAE) to consider the uncertainty of user preferences for regularized representation of historical trajectories~\cite{gao2019predicting}. A meta-learning technique called METAODE also employed variational Bayes to encode past human movement patterns into latent space~\cite{tan2021meta}. In essence, these approaches principally rely on integrating numerous semantics including sequential information into a unified space while omitting the possibility of disentangling it to expose the characteristics of human mobility patterns.

\subsection{Disentanglement Learning}

The privilege of disentanglement learning is that it enables an interpretable perspective to understand the multiple inherent motions/factors behind the intricate data representations in addition to notable expressiveness. To disentangle the learned representations, most recent studies developed VAEs such as $\beta$-VAE to optimize the mutual interaction between different latent factors~\cite{burgess2018understanding,chen2018isolating,bai2021contrastively}. For example, $\beta$-VAE~\cite{burgess2018understanding} is a simple but effective variant of the ordinary VAE that severely penalizes the Kullback–Leibler (KL) divergence term for disentanglement learning. Li et al. presented a Disentangled Sequential Autoencoder (DSVAE) approach for sequential data (e.g., video), aiming at factorizing the latent variables into static and dynamic parts~\cite{li2018disentangled}. To make the latent variables interpretable and controllable, a latent variable guidance-based generative model called Guided-VAE makes an effort to utilize VAE to learn a transparent representation~\cite{ding2020guided}.
Bai et al. presents a sequential VAE to learn disentangled representations in a self-supervised manner~\cite{zhu2020s3vae}. Bai et al. also extend the sequential VAE with a self-supervised learning approach to facilitate the factorization of video representations~\cite{bai2021contrastively}. In addition, The newly developed self-supervised learning offers a new avenue to drive the acquisition of semantic representations~\cite{huang2022self}. For example, Ma et al. employ the ideas of latent self-supervision and intention disentanglement to boost the convergence of representation learning and utilize it in sequential recommendation tasks~\cite{ma2020disentangled}. In sum, the success of these approaches suggests that, in addition to facilitating the understanding of rich semantics underlying data, disentangling the representation into distinct parts can make the representation more transparent and interpretable.

\section{Preliminaries}
\label{3_Preminilaries}
%We first present the essential definitions and formulate the next POI prediction problem. The relevant background details for VAE and contrastive estimation are then provided. %Additionally, Table XX presents the frequently used notations throughout this paper.

\subsection{Problem Definition}

\textbf{Definition 1 (POI). }
Let $l \in \mathcal{L}$ denotes a POI tagged by the location-based systems, and each POI corresponds to a geographic coordinate (e.g., longitude $lo$ and latitude $la$) and a category $ca$ (e.g., restaurant, museum, or park).

\noindent
\textbf{Definition 2 (Check-in Sequence). }
A check-in sequence (or trajectory) $T_u=\{l_{1}^u,l_{2}^u,\cdots,l_{n}^u\}$ left by user $u$ is a sequence of $n$ POIs ordered by visiting time, where $l_{\tau}^u$ means a user $u$ visit POI $l$ at time $t_\tau$ ($\tau \in \{1,2,\cdots,n\}$). Let $\mathcal{T}_u=\{T_u^1,T_u^2,\cdots,T_u^{m}\}$ denote $m$ historical trajectories of user $u$, where each trajectory $T_u^i$ contains a sequence of POIs ordered by visiting time, e.g., $T_u^i=\{l_{1}^{i,u},l_{2}^{i,u},\cdots, l_{n}^{i,u}\}$.

Formally, given a user $u$ with his/her recently visited check-in sequence $T_u^m=\{l_{1}^{m,u},l_{2}^{m,u},\cdots,l_{n}^{m,u}\}$ and entire historical trajectory $\mathcal{T}_u$ , our goal is to predict a POI $l_{n+1}^{m,u}$ for user $u$ to visit next. Notably, we mainly target disentangled representation learning for users' recently visited POI sequences. For simplicity, we will omit user identity (i.e. $u$) and trajectory index (i.e. $m$) in the following sections.

\subsection{Variational Bayes}

Variational Autoencoder (VAE)~\cite{kingma2013auto} containing an encoder and a decoder operates the input data $\boldsymbol{x}$ into a latent space, where the latent variables are denoted by $\boldsymbol{z}$. Thus, the marginal likelihood $\log p(\boldsymbol{x})$ can be obtained by maximizing the Evidence Lower BOund (ELBO), which is defined as:
\begin{align}
    \label{eq:vae}
    &\log p_\theta(\boldsymbol{x}) \geq \\\nonumber
    &\mathbb{E}_{\boldsymbol{z}\sim q_\phi(\boldsymbol{z}|\boldsymbol{x})}[\log p_\theta(\boldsymbol{x}|\boldsymbol{z})]-\mathit{KL}[q_\phi(\boldsymbol{z}|\boldsymbol{x}||p(\boldsymbol{z}))].
\end{align}
Herein, $q_\phi(\boldsymbol{z}|\boldsymbol{x})$ is an approximate posterior distribution, parameterized by $\phi$, $p_\theta(\boldsymbol{x}|\boldsymbol{z})$ with parameters $\theta$ is a likelihood function, and $p(\boldsymbol{z})$ is a prior (e.g., Gaussian prior) over the latent variables.

\subsection{Contrastive Estimation}
In recent self-supervised learning paradigms~\cite{tian2020makes,zhou2021self}, mutual information (MI) is a common measure of the mutual dependence or compatibility between two variables. 
Specifically, they usually employ the noise contrastive estimation (NCE)~\cite{gutmann2010noise,oord2018representation} to maximize the lower bound on the mutual information, which can be denoted as follows:
\begin{align}
    \label{obj-NCE}
    \mathcal{L}_{\mathit{NCE}}&=\\\nonumber
    &\mathbb{E}\left[-\log \left( \frac{\exp^{g(x)^\top g(x^+)}}{\exp^{g(x)^\top g(x^+)}+\sum_{j=1}^{J} \exp^{g(x)^\top g(x^-)}}\right)\right],
\end{align}
where $x$, $x^+$, and $x^-$ respectively denote the \textit{anchor}, \textit{positive}, and \textit{negative} instances. Besides, $ \exp^{g(\cdot)^\top g(\cdot)} $ is a similarity measure (e.g., cosine similarity) between two instances.

\section{Architecture Design}%Methodology
\label{Methodology}
\begin{figure*}[t]
    \centering
    \includegraphics[width=0.95\textwidth]{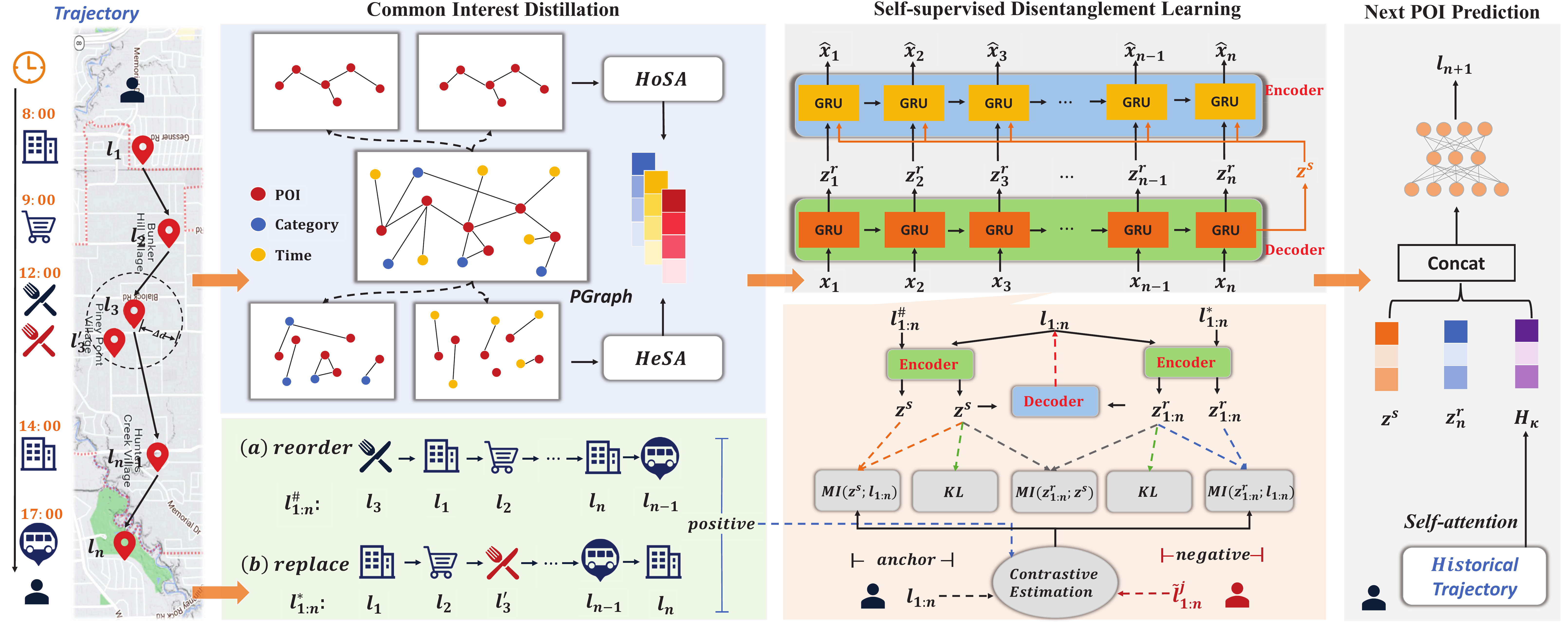}
    \caption{The pipeline of proposed SSDL.}
    \label{fig:dismove}
\end{figure*}
%We first overview the pipeline of our proposed DisMove, followed by the detail of each significant component. Also, the details of optimization and inference are presented.
%\subsection{The Overall Architecture}
We make an overview of our proposed framework SSDL in Fig.~\ref{fig:dismove}, which mainly comprises three components. First, we build a POI-centric Graph (PGraph) to explore the common interests from the entire user trajectories and make interest aggregation to obtain both homogeneous and heterogeneous semantics underlying each POI. Then, our Self-supervised Disentanglement Learning component attempts to produce the time-invariant and time-varying variables for each trajectory. At last, SSDL uses the disentangled representations as well as the user's long-term preference modeled by an attentive network to predict the next POI.

\subsection{Common Interest Distillation}

To distill multiple correlations behind the POIs and their affiliated context, we build a POI-centric graph (PGraph).

\subsubsection{Graph Structure and Building Process}

Incorporating prior correlations and multiple common interests are critical to obtain a good POI representation and understand human diverse mobility patterns. 
As several elements are recorded by LBSN, e.g., POI identity, geographical coordinate, visiting time, and POI category, we concentrate on exploring four contextual semantics to build our PGraph, including \textit{consecutive}, \textit{geospatial}, \textit{time-aspect}, and \textit{activity-aspect} interests.

Let $\mathcal{G}=(\mathcal{V},E)$ denotes our PGraph that models the human common interests, where $\mathcal{V}=(\mathcal{V}_l\cup \mathcal{V}_t \cup \mathcal{V}_a)$ is the set of nodes, and $E=(E_{c} \cup E_{g} \cup E_{t} \cup E_{a})$ is the set of edges. Here $\mathcal{V}_l = \mathcal{L}$ represents a collection of different POIs, $\mathcal{V}_t$ is the set of time bins, $\mathcal{V}_a$ denotes the set of POI categories, and $E_{c}, E_{g}, E_{t}, E_{a}$ indicate the above four contextual semantics, respectively. That is to say, $\mathcal{G}$ contains four sub-graphs, each of which represents an important user interest. 
Four sub-graphs are described in the following four paragraphs.

\noindent\textbf{Consecutive Interest.}
According to \cite{gao2017identifying}, among millions of POIs in location-based systems, (1) people typically visit only a small subset of POIs that appeal to them; and (2) some POIs are visited more frequently than others. 
This phenomenon demonstrates that human mobility contains some common transitional regularities behind their past check-ins. Therefore, we consider that it is necessary to capture the consecutive correlations between distinct POIs to reveal human motion-based interests. Correspondingly, we formulate a weighted sub-graph $\mathcal{G}_c=(\mathcal{V}_l,{E}_c, \boldsymbol{A}_c)$ to describe such diverse correlations, where $\mathcal{V}_l$ is the set of distinct POIs, ${E}_c$ is the edge set, and $\boldsymbol{A}_c \in \mathbb{R}^{|\mathcal{L}|\times |\mathcal{L}|}$ refers to the adjacency matrix. Given two POIs (e.g., POI $l_i$ and POI $l_j$) that are successively visited, we create an edge between them and then calculate the edge weight (i.e., entry $
A_c^{ij} \in \boldsymbol{A}_c$) using the corresponding transitional probability. Formally, such an edge weight can be defined as:
\begin{equation}
\label{eq:freq-c}
A_{c}^{i j}=f_{c}^{i j} / f_{c}^{i},
\end{equation}
where $f_c^{ij}$ refers to the frequency of edge $l_i\rightarrow l_j$ appeared in the check-in data, and $f_c^{i}$ denotes the frequency of POI $l_i$ appeared in the check-in data. As such, we are able to acquire the matrix $\boldsymbol{A}_c$ to preserve the consecutive interests underlying the trajectories.

\noindent\textbf{Geographical Interest.}
People are more likely to visit nearby POIs than distant ones \cite{gao2022contextual}. 
Motivated by this, we formulate an undirected sub-graph $\mathcal{G}_g=(\mathcal{V}_l,E_g, \boldsymbol{A}_g)$ to describe such interactions, where $E_g$ is the set of edges and $\boldsymbol{A}_g$ ($\in \mathbb{R}^{|\mathcal{L}|\times |\mathcal{L}|}$) denotes the adjacency matrix regarding geographical interests. Given POI $l_i$ and $l_j$, the edge weight $A_g^{ij}$ ($\in \boldsymbol{A}_g$) can be calculated as:
\begin{equation}
A_g^{ij}=\left\{\begin{array}{l}
0, \quad \operatorname{g}\left(l_{i}, l_{j}\right)>\Delta g; \\
1, \text{otherwise}.
\end{array}\right.
\end{equation}
Herein, $g(l_{i}, l_{j})$ is the great-circle distance function, $\Delta g$ is a predefined threshold to restrict the impact of geographical noise. In this paper, we set $\Delta g=3\ km$. 

\noindent\textbf{Time--aspect Interest.}
For each check-in, it is associated with a visiting timestamp, reflecting the human temporal semantics. 
As it is a key factor for understanding human periodic regularity, we propose to investigate the mutual interactions between POI and visiting time to obtain the time-aspect interest. However, each visiting timestamp is actually a continuous value, we follow previous studies and aggregate all of the visiting timestamps into the hour-level time bins~\cite{jeon2021lightmove,chen2021curriculum}. Meanwhile, people may respectively show different preferences on weekday and weekend, we thus assign 48 time bins to replace the original visiting timestamps, where the weekday and weekend are specified. %Let $$
We thus formulate a weighted sub-graph $\mathcal{G}_t (\mathcal{V}_l \cup \mathcal{V}_t, E_t, \boldsymbol{A}_t)$, where $\boldsymbol{A}_t$ maintains human time-aspect interest. Similar to the above graph $\mathcal{G}_c$, we can also calculate the time-aspect interest between the POI $l_i$ and time bin $t_\tau$ by:
\begin{equation}
A_{t}^{i \tau}=f_{t}^{i \tau} / f_{t}^{i},
\end{equation}
where $f_{t}^{i \tau}$ denotes the frequency of visiting POI $l_i$ at time $t_\tau$, and $f_{t}^{i}$ is the total number that POI $l_i$ has been visited.

\noindent\textbf{Activity-aspect Interest.}
A user who wants to post a check-in to LBSNs indicates that he/she is engaged in a specific type of activity that appeals to him/her. In practice, each POI has a contextual description (i.e., POI category) that reflects a real-world activity, we consider that taking into account such contextual interactions is an essential addition to understanding human preferences. Notably, the number of POI categories is much smaller than the number of POIs. As a result, linking a POI to its category can offer a coarse-grained perspective on the higher-order interactions between various POIs. To this end, we build an undirected graph $\mathcal{G}_a (\mathcal{V}_l \cup \mathcal{V}_a, E_t, \boldsymbol{A}_a) $ to describe the activity-aspect interest. To be more precise, we explicitly build an edge between a POI and the contextual category it belongs to, and then we treat each category as a regular node in $\mathcal{G}_a$. %More concretely

\subsubsection{Interest Aggregation}
To extract the semnatic contexts underlying POIs from the PGraph, we propose to adopt graph neural networks (GNNs) which have been widely applied in numerous graph-based tasks and obtained remarkable success.

\noindent\textbf{Homogeneous Semantic Aggregation (HoSA).}
According to the structure of the built PGraph, we can find that consecutive interest and geospatial interest that belong to the homogeneous semantics as they only contain the nodes of POI identities. Thus, HoSA attempts to aggregate the underlying information from the nodes of the same type, i.e., POI identity. First, the consecutive correlation matrix $\boldsymbol{A}_c$ reflects human real-world transitional preferences, we can naturally regard each POI's transitional distribution as its prior feature to describe the relationship between a specific POI and its neighbors. To this end, we set each $\boldsymbol{A}_c^i$ as the initial feature of the POI node $\mathcal{V}_p^i$. Besides, the geospatial correlation matrix $\boldsymbol{A}_g$ preserves the geographical closeness between different POIs, providing the weak signal of human potential transitional tendencies. Hence, $\boldsymbol{A}_g$ can be regard as an augmentation of the consecutive correlation matrix $\boldsymbol{A}_c$. Therefore, we merge these two matrices into a unified matrix $\boldsymbol{A}_h$ to reveal observed and unobserved preferences of transitional dependencies. Specifically, given two distinct POI nodes $\mathcal{V}_l^i$ and $\mathcal{V}_l^j$, its correlation score $A_h^{ij}$ is defined as:
\begin{equation}
    A_h^{ij}=\left\{\begin{array}{l}
A_{c}^{i j}, \text { if } A_{c}^{i j} \neq 0;\\
A_{g}^{i j}, \text { others } .
\end{array}\right.
\end{equation}
For any POI node $\mathcal{V}_l^i$, we embed each POI node to a unified representation:
\begin{equation}
    \boldsymbol{s}_i^l=\boldsymbol{A}_h^{i}\boldsymbol{W}_l+\boldsymbol{b}_l,
\end{equation}
where $\boldsymbol{W}_l \in \mathbb{R}^{|\mathcal{L}|\times d}$ and $\boldsymbol{b}_l \in \mathbb{R}^{d}$ are trainable matrices. The dimension of $\boldsymbol{s}_i^l$ is $d$. Afterwards, each POI has its unique initial representation. To bridge the correlation between POI $\mathcal{V}_l^i$ and each of its neighbor $\mathcal{V}_l^j\in \Omega(\mathcal{V}_l^i)$, we devise a scoring function to evaluate the different contributions of neighboring nodes. For instance, given POI node $\mathcal{V}_l^i$ and its neighbor $\mathcal{V}_l^j$, we define contribution measure as:
\begin{equation}
    a(\boldsymbol{s}_i^l,\boldsymbol{s}_j^l)=\boldsymbol{b}_a^{T}[\boldsymbol{s}_i^l \oplus \boldsymbol{s}_j^l],
\end{equation}
where $\oplus$ is the concatenation operation and $\boldsymbol{b}_a \in \mathbb{R}^{2d}$ is a learnable vector. Then, we follow the standard GAT~\cite{vel2018graph} and use softmax function to normalize the attention scores across all neighbors of POI $\mathcal{V}_l^i$, where each attention score regarding its neighbour $\mathcal{V}_l^j$ can be formulated as:
\begin{equation}
\alpha_{i j}=\frac{\exp (\text{LeakyReLU}(a(\boldsymbol{s}_i^l,\boldsymbol{s}_j^l)))}{\sum_{k \in \Omega(\mathcal{V}_l^i)} \exp (\text{LeakyReLU} (a(\boldsymbol{s}_i^l,\boldsymbol{s}_k^l)))}.
\end{equation}
In the end, we obtain the aggregated representation $\boldsymbol{e}^l_i$ of POI node $\mathcal{V}_l^i$ by a sum operation:
\begin{equation}
\boldsymbol{e}_{i}^l=\sigma(\sum_{j \in \Omega(\mathcal{V}_l^i)} \alpha_{i j}  \boldsymbol{s}_{j}^l\boldsymbol{W}_e),
\end{equation}
where $\sigma$ is the sigmoid activation function and $\boldsymbol{W}_e \in \mathbb{R}^{d\times d}$ is a set of trainable parameters.

\noindent\textbf{Heterogeneous Semantic Aggregation (HeSA).}
HeSA is to aggregate the associated information of POIs from the neighboring nodes with different types. In our PGraph, there are two correlations that describe the heterogeneous semantics between different types of nodes, i.e., the time-aspect and activity-aspect interests. In contrast to HoSA, we do not involve the attention mechanism to quantify the different contributions of POI's heterogeneous neighbors. The reason is that the number of them are extremely smaller than that of the POIs, we thus attempt to capture all of the possible heterogeneous neighbors of a given POI directly to enhance the semantic information. 

\noindent(1) For time–aspect interest, each POI $p_i$ is associated with a probability distribution $\boldsymbol{A}_t^i$ ($\in \boldsymbol{A}_t$) that describes the preference strengths between POI and time bins. We leverage the message-passing neural network inspired by~\cite{rao2022graph} to incorporate the time-aspect preference of each POI, which can be formulated as:
\begin{equation}
\boldsymbol{e}_i^{t}=\text{tanh}\left(\boldsymbol{A}_{t}^i \mathbf{W}_{t}\right),
\end{equation}
where tanh is the activation function and $\mathbf{W}_t$ is a trainable matrix. Finally, we can obtain each POI's temporal context.

\noindent(2) For activity-aspect interest, we obtain each POI's activity-aware semantic by:

\begin{equation}
\boldsymbol{e}_i^{a}=\text{tanh} \left(\boldsymbol{A}_{a}^i \mathbf{W}_{a}\right),
\end{equation}
where $\mathbf{W}_a$ is a trainable matrix. Finally, the homogeneous and heterogeneous semantics behind each POI are acquired by HoSA and HeSA, respectively. In the following mobility encoding procedures, we will use these contextual representations as the embeddings of POIs in user trajectories. And these embeddings can be jointly optimized during self-supervised learning and task learning.

\subsection{Context-aware Mobility Encoding} 
Existing studies usually choose the recurrent neural networks such as Long-short Term Memory (LSTM) or Gated Recurrent Unit (GRU) to capture human transitional regularities. Since the complex stacked gate operations in LSTM typically struggle with the gradient vanishing problem, we select GRU as the kernel of our mobility encoder. For each 
$l_\tau$ in a given trajectory $T=\{l_{1},l_{2},\cdots,l_{n}\}$, we have collected the homogeneous and heterogeneous semantics behind it. In this way, they can be viewed as reflections of different interest in different domains. Therefore, we extend the GRU cell to capture the sequential information as well as the contextual information behind each POI. Correspondingly, the recursive process with GRU can be formulated as follows: 
\begin{align}
    \label{eq:emd}
    &     \boldsymbol{c}_\tau=[\boldsymbol{e}_\tau^l \oplus \boldsymbol{e}_\tau^t \oplus \boldsymbol{e}_\tau^a] \boldsymbol{W}_f+\boldsymbol{b}_f,\\
    & \boldsymbol{h}_\tau=\text{GRU}(\boldsymbol{c}_\tau,\boldsymbol{h}_{\tau-1}),
    \label{gru}
\end{align}
where $\boldsymbol{h}_\tau$ and $\boldsymbol{h}_{\tau-1}$ are the hidden states of the current POI $l_\tau$ and the last visited POI $l_{\tau-1}$, respectively. Herein, $\boldsymbol{c}_\tau$ is the contextual embedding of POI $l_\tau$, which is a unified representation that integrates the homogeneous and heterogeneous semantics of POI $l_\tau$ (they include $\boldsymbol{e}_\tau^l$, $\boldsymbol{e}_\tau^t$, and $\boldsymbol{e}_\tau^a$). In addition, $\boldsymbol{W}_f$ and $\boldsymbol{b}_f$ are trainable parameters.

\subsection{Self-supervised Disentanglement Learning}
Now we describe in detail the self-supervised disentanglement learning in SSDL.

\subsubsection{Disentanglement via Variational Inference}
Given any recent trajectory $T=\{l_1,l_2,\cdots,l_n\}$, we attempt to learn a set of time-varying variables $z_{1:n}^r=\{z_{1}^r,z_{2}^r,\cdots,z_{n}^r\}$ and a time-invariant variable $z^s$, where $z_{1:n}^r$ aims at exploring the dynamics of human time-dependent interests while $z^s$ undertakes the role of learning human inherent time-independent periodicity (habits). Formally, let $z_\tau$ be the entangled latent code of check-in $l_\tau$, and we have $z_\tau=(z_\tau^r,z^s)$. For consistency, let $l_{1:n}$ denote the check-in sequence $\{l_1,l_2,\cdots,l_n\}$. %
As people's future movements are affected by their previous check-in behaviors, we assume that each $z_\tau$ depends on its previous states $z_{<\tau}=\{z_1,z_2,\cdots,z_{\tau-1}\}$. 
In addition, as user's long-standing interests will not be changed dramatically by recent activities, we assume that 
$z_{1:n}^r$ and $z^s$ are independent from each other, i.e., $p(z_{1:n})=p(z_{1:n}^r)p(z^s)$. Hence, we formulate our probabilistic generative model as follows:
\begin{align}\textbf{Prior: }
    p(l_{1:n},z_{1:n})&=p(z_{1:n})p(l_{1:n}|z_{1:n})\\\nonumber
    &=[p(z^s)\prod_{\tau=1}^n p(z_\tau^r|z_{<\tau}^r)]\cdot \prod_{\tau=1}^n p(l_\tau|z_\tau^r,z^s),
\end{align} 
where $p(z_{1:n})$ is a prior. Herein, we choose the Gaussian distribution as the prior $p(z^s)$, i.e., $p(z^s) \sim \mathcal{N}(\mathbf{0},\mathbf{1})$. We follow the rule of standard variational Bayes and set $\mathcal{N}(\mu(z_{<\tau}),\sigma^2(z_{<\tau}))$ as $p(z_\tau^r|z_{<\tau}^r)$, where $\mu(\cdot)$ and $\sigma^2(\cdot)$ can be modeled by popular recursive networks. In practice, we also use GRU to obtain $z_\tau^r$ as follows:
\begin{align}
\label{eq:z_r}
    \boldsymbol{h}_\tau^r=\text{GRU}(\boldsymbol{h}_\tau,z_{\tau-1}^r),
    z_\tau^r=\mu(\boldsymbol{h}_\tau^r)+ \sigma(\boldsymbol{h}_\tau^r)\odot \epsilon,
\end{align}
where $\epsilon \sim \mathcal{N}(\mathbf{0},\mathbf{1})$ and $\odot$ is element-wise multiplication.

Subsequently, we expect to produce a posterior distribution $q(z_{1:n}|l_{1:n})$ to cater to the learning manner of variational inference. Thus, we define the posterior as follows:
\begin{align}\textbf{Posterior: }
q(z_{1:n}|l_{1:n})&=q\left(z^s, z_{1: n}^r \mid l_{1: n}\right)\\\nonumber
&=q(z_{1:n}^r|l_{1:n})q(z^s|l_{1:n})\\\nonumber
&=q\left(z^s \mid l_{1: n}\right) \prod_{\tau=1}^{n} q\left(z_\tau^r \mid z_{<\tau}^r, l_{\leq\tau}\right).
\end{align}
We note that the above process is also operated in an auto-regressive manner.
We use another GRU cell that has the same architecture as the Mobility Encoding network to generate posterior distributions. 
At last, we obtain the Evidence Lower BOund (ELBO) as follows:
\begin{align}
\textbf{ELBO:}\ &\max _{p, q} \mathbb{E}_{l_{1: n} \sim p_{D}} \mathbb{E}_{q\left(z_{1:n} \mid l_{1: n}\right)}[\log p\left(l_{1: n} \mid z_{1:n}\right)\\\nonumber
&-K L\left[q\left(z_{1:n} \mid l_{1: n}\right) \| p(z_{1:n})\right]],
\end{align}
where $p_D$ is the empirical trajectory distribution. As $z_{1:n}$ is comprised of mutually independent $z^s$ and $z_{1:n}^r$, the second term of KL-divergence can be disentangled as:
\begin{align}
&\mathit{KL}\left[q\left(z_{1:n} \mid l_{1: n}\right) \| p(z_{1:n})\right]=\\\nonumber
&K L\left[q\left(z^s \mid l_{1: n}\right) \| p(z^s)\right]+K L\left[q\left(z_{1: n}^r \mid l_{1: n}\right)|| p\left(z_{1: n}^r\right)\right].
\end{align}
Following the principle of VAE~\cite{kingma2013auto,bai2021contrastively}, we present a theoretical proof to illustrate above modeling processes. %That is:
\begin{proof}
\begin{equation}
\label{eq:proof}
\begin{aligned}
& \log p\left(l_{1: n}\right) \\
\geq &-KL\left[q\left(z_{1:n} \mid l_{1: n}\right) \| p\left(z_{1:n} \mid l_{1: n}\right)\right]+\log p\left(l_{1: n}\right) \\
=&-KL\left[q\left(z^s, z_{1: n}^r \mid l_{1: n}\right) \| p\left(z^s, z_{1: n}^r \mid l_{1: n}\right)\right]+\log p\left(l_{1: n}\right) \\
=& \mathbb{E}_{q\left(z^s, z_{1: n}^r \mid l_{1: n}\right)}[\log p\left(l_{1: n} \mid z^s, z_{1: n}^r\right)\\
&-\log q\left(z^s, z_{1: n}^r \mid l_{1: n}\right)+\log p\left(l_{1: n}\right)] \\
=& \mathbb{E}_{q\left(z^s, z_{1: n} \mid l_{1: n}\right)}[\log p\left(l_{1: n} \mid z^s, z_{1: n}^r\right)\\
&-\log q\left(z^s, z_{1: n}^r \mid l_{1: n}\right)+\log p\left(z^s, z_{1: n}^r\right)] \\
=& \mathbb{E}_{q\left(z^s, z_{1: n}^r \mid l_{1: n}\right)}[\log p\left(l_{1: n} \mid z^s, z_{1: n}^r\right)-\log q\left(z^s \mid l_{1: n}\right)\\
&-\log p\left(z_{1: n}^r \mid l_{1: n}\right)+\log p(z^s)+\log p\left(z_{1: n}^r\right)] \\
=& \mathbb{E}_{q\left(z^s, z_{1: n}^r \mid l_{1: n}\right)}\left[\log p\left(l_{1: n} \mid z^s, z_{1: n}^r\right)\right]\\
&-K L\left[q\left(z^s \mid l_{1: n}\right) \| p(z^s)\right]\\
&-K L\left[q\left(z_{1: n}^r \mid l_{1: n}\right)|| p\left(z_{1: n}^r\right)\right].
\end{aligned}
\end{equation}
\end{proof}
Recall that the results of the above proof are similar to the results in standard VAE, which usually confront the agnostic prior distribution that causes posterior collapse problem and leaves the learned latent space still entangled~\cite{akuzawa2021information}. 
This phenomenon has been revealed in recent studies, e.g., $\beta$-VAE~\cite{burgess2018understanding} and $\beta$-TCVAE~\cite{chen2018isolating}. Additionally, \cite{locatello2019challenging} provides us with a clearer perspective that reveals the challenges with disentangled representation in variational inference. Thus, we conjecture that the last two terms regularized by KL-divergence in Eq.(\ref{eq:proof}) are hard to close to their corresponding prior, which would make each posterior become non-informative. For the purpose of receiving clean disentanglement of $z^s$ and $z_{1:n}^r$, we are inspired by recent self-supervised learning and enforce disentanglement mobility learning from the perspective of mutual information.
\subsubsection{Mutual Information Regularization}
We now turn to detail on how to combine contrastive learning with disentangled mobility learning. We first introduce variational mobility learning from the perspective of Mutual Information (MI). The goal of MI is a measure of the mutual dependence between two variables. Since both $z^s$ and $z^r_{1:n}$ are derived from the original trajectory, we thus add three additional MI terms to regularize the latent space of them, which can be defined as follows:
\begin{align}
\label{eq:obj}
&\mathcal{J}_{self}=\max _{p, q} \mathbb{E}_{l_{1: n} \sim p_{D}} \mathbb{E}_{q\left(z_{1:n} \mid l_{1: n}\right)}[\log p\left(l_{1: n} \mid z_{1:n}\right)\\\nonumber
&-\alpha (K L\left[q\left(z^s \mid l_{1: n}\right) \| p(z^s)\right]+K L\left[q\left(z_{1: n}^r \mid l_{1: n}\right)|| p\left(z_{1: n}^r\right)\right])\\\nonumber
&+\beta (MI\left(z^s ; l_{1: n}\right)+MI\left(z_{1: n}^r ; l_{1: n}\right))-\gamma MI\left(z_{1: n}^r ; z^s\right),
\end{align}
where $\alpha$, $\beta$ and $\gamma$ are weight coefficients. $MI(\cdot,\cdot)$ refers to MI term. For instance, $MI\left(z^s ; l_{1: n}\right)$ is defined as:
\begin{align}
    \mathbb{E}_{q(z^s,l_{1:n})}\left[\log \frac{q(z^s|l_{1:n})}{q(z^s)}\right].
\end{align}
We note that other MI terms have the similar formulation. Now our goal become enforcing the posteriors matching with their corresponding priors while ensuring that $z^s$ and $z^r_{1:n}$ are disentangled from each other. Note that the complete proof of Eq.(\ref{eq:obj}) is provided in Appendix part. To estimate the MI terms, we follow most of recent studies~\cite{gutmann2010noise,tian2020makes,zhou2021self} and employ the NCE loss to make contrastive estimation. For instance, a contrastive estimation of $MI(z^s;l_{1:n})$ can be defined as follows,
\begin{align}
\label{nce:zs}
&\mathcal{C}_{z^s}
\approx \mathbb{E}_{p_{D}} \log \frac{\psi\left(z^s, l_{1: n}^{+}\right)}{\psi\left(z^s, l_{1: n}^{+}\right)+\sum_{j=1}^{m} \psi\left(z^s, \Tilde{l}_{1: n}^{j}\right)},%+\log (n+1)
\end{align}
where $\psi(\cdot, \cdot)=exp(sim(\cdot, \cdot)/\eta)$, $sim(\cdot, \cdot)$ denotes the cosine similarity function, $m$ is the number of negative trajectories, and $\eta=0.5$ is a temperature parameter.
Notably, we treat $l_{1:n}$ as the \textit{positive} trajectory sequence regarding $z^s$ and specify it using $l_{1:n}^{+}$. Besides, $\Tilde{l}_{1: n}^{j}$ refers to a \textit{negative} sample (trajectory), which is generated from other users. 

\textbf{Augmentation for time-invariant factor.} However, due to the limited scale of \textit{positive} samples, we try to generate more realistic trajectories to augment the original samples. As $z^s$ reveals human intrinsic periodicity and should not be affected by recent moving behaviors, i.e., time-independent, we can thus randomly change the order of a given trajectory $l_{1:n}$ and formulate several augmentation versions w.r.t $l_{1:n}$. We claim that it is a simple but efficient strategy to obtain rich augmented samples. Correspondingly, the contrastive estimation regarding these augmented samples can be denoted as follows:
\begin{align}
&\mathcal{C}_{z^s}
\approx \mathbb{E}_{p_{D}} \log \frac{\psi\left(z^s, l_{1: n}^{\#}\right)}{\psi\left(z^s, l_{1: n}^{\#}\right)+\sum_{j=1}^{m} \psi\left(z^s, l_{1: n}^{j}\right)},%+\log (n+1)
\end{align}
where $l_{1:n}^{\#}$ indicates it is an augmentation version of $l_{1:n}$. For time-invariant factor $z^s$, we can use the collected samples including augmented samples to make a final estimates as follows:
\begin{align}
\label{eq:mi_zs}
    MI\left(z^s ; l_{1: n}\right)\approx \frac{1}{2} (\mathcal{C}_{z^s}+\mathcal{C}^{\#}_{z^s}).
\end{align}
\textbf{Augmentation for time-varying factor.} As for $z^r_{1:n}$ is a set of latent variables regarding $l_{1:n}$, showing human human time-dependent interests. Similar to Eq.~(\ref{nce:zs}), we can obtain the contrastive estimation of $MI(z^r_{1:n};l_{1:n})$ as $\mathcal{C}_{z_{1: n}^r}$. Furthermore, we provide another data augmentation method to enhance the optimization of $MI(z^r_{1:n};l_{1:n})$. The intuition is that $z^r_{1:n}$ is a set of time-dependent variables. In practice, real-world check-in data could be subject to noise and uncertainty due to the presence of collective POIs~\cite{sun2021point}. Hence, a user usually posts an fuzzy POI to replace her accurate position, which could weaken human mobility pattern learning and even result in inaccurate predictions. Motivated by~\cite{zhang2021interactive,sun2021point}, it is encouraging that we can use any member of related collective POI to replace the original POI in a trajectory to obtain an augmentation trajectory for time-varying factor training, which will not change any temporal semantics. In addition, another potential benefit of such a practice is to alleviate the uncertainty issue behind diverse human check-in behaviors. In our implementation, we use neighbors of the same category within 300m of a given POI as members of its collective POI and replace about 30\% of the POIs in a given trajectory with their related collective POIs, which does not heavily affect the full semantics behind the original trajectory. As a result, we can obtain a large number of synthetic trajectories that provide multiple views of a given trajectory. Similar to Eq.(\ref{eq:mi_zs}), we can get the final estimation regarding $z_{1:n}^r$ as follows:
\begin{align}
\label{eq:mi_zr}
    MI\left(z^r_{1:n} ; l_{1: n}\right)\approx \frac{1}{2} (\mathcal{C}_{z^r_{1:n}}+\mathcal{C}^*_{z^r_{1:n}}),
\end{align}
where $\mathcal{C}^*_{z^r_{1:n}}$ is the contrastive estimation of the augmented trajectories regarding time-varying factors. As for the final term $MI(z_{1: n}^r;z^s)$, the variables in it are all in the latent space, we thus can directly choose the standard mini-batch weighted sampling (MWS)~\cite{chen2018isolating} for comparative estimation.

\subsection{Task Learning}
So far, we have obtained a set of time-varying variables and a time-invariant variable for each trajectory. We turn to use our task learning network to predict the next POI. For each user, we actually own her entire historical trajectory. Inspired by~\cite{feng2018deepmove,gao2019predicting} modeling such a long trajectory would boost the capture of human long-term transitional preferences. It is natural to adopt the RNN to encode the transitional regularity underlying human historical trajectory. But in practice, there are massive time-ordered POIs in their historical trajectories, which usually result in a serious time cost problem. Therefore, we employ a self-attention layer with position encoding to capture the taste of the transitional behavior of a user as well as the long-distance dependencies. Given a user's entire historical trajectory $\mathcal{T}_{1:\mathcal{K}}$ containing $K$ ordered POIs. We first reuse the linear layer (cf. Eq.\ref{eq:emd}) to obtain the dense representation of each POI in $\mathcal{T}_{1:\mathcal{K}}$. Correspondingly, we use $\boldsymbol{\mathcal{T}}_{1:\mathcal{K}}$ to denote the trajectory with embedded POIs. To determine the order of POIs in $\boldsymbol{\mathcal{T}}_{1:\mathcal{K}}$, we follow~\cite{vaswani2017attention} and use the sine/cosine function-based position embedding to formulate the final representation of each POI, which can be denoted as:
\begin{align}
\label{eq:pos}
    \boldsymbol{\mathcal{T}}'_i=\boldsymbol{\mathcal{T}}_i+\Phi(\boldsymbol{\mathcal{T}}_i),
\end{align}
where $\Phi(\boldsymbol{\mathcal{T}}_i)$ is the position embedding of POI $l_i$ in $\mathcal{T}_{1:\mathcal{K}}$. Then, we employ one-layer self-attention network to receive a set of hidden states regarding $\boldsymbol{\mathcal{T}}$, as follows:
\begin{align}
    \label{eq:his}
    \boldsymbol{H}_{1:\mathcal{K}}=\text{self-att}(\boldsymbol{\mathcal{T}}'_{1:\mathcal{K}}).
\end{align}
In our study, we use the last state $\boldsymbol{H}_\mathcal{K}$ to represent $\mathcal{T}_{1:\mathcal{K}}$ and regard it as one of the inputs for task prediction.

Now we take $z^r_{1:n}$, $z^s$, and $\boldsymbol{H}_{1:\mathcal{K}}$ as the input and employ a one-layer fully-connected network with softmax function to obtain the predict POI. the process can be expressed as:
\begin{align}
\label{eq:pred}
    \Tilde{l}_{n+1}=\arg \max(\text{softmax}([z^r_{n} \oplus z^s \oplus \boldsymbol{H}_{\mathcal{K}}]\boldsymbol{W}_t+\boldsymbol{b}_t))
\end{align}
Correspondingly, the loss function for trajectory $T$ can be expressed as:
\begin{align}
\label{eq:loss}
    \mathcal{L}_T=-l_{n+1}\log \Tilde{l}_{n+1}.
\end{align}
To minimize the above cross-entropy loss, we employ Adam algorithm to optimize the parameters. We outline the complete pipeline of training SSDL in Algorithm~\ref{Algo-ST}.
\begin{algorithm}[ht]
    \footnotesize
	\caption{The pipeline of training SSDL.}
	\label{Algo-ST}
	\BlankLine
	
	\KwIn{POI set $\mathcal{L}$; Historical trajectories $\mathcal{T}$ and current trajectory $T$ of users. }
\tcc{Common Interest Distillation}
Build the PGraph from entire trajectory data;\\
Generate the homogeneous and heterogeneous semantics via HoSA and HeSA for each POI;\\
\tcc{Disentanglement learning}
	\Repeat{convergence}{
	\ForEach{$T$}
	{
	 Compute each POI embedding via Eq.~(\ref{eq:emd});\\
	Obtain each hidden state $\boldsymbol{h}_\tau$ via Eq.(\ref{gru});\\
	Compute each $z_\tau^r$ via Eq.(\ref{eq:z_r});\\
	Compute $z^s$ based on the last hidden state;\\
	Make trajectory augmentation regarding $z^s$;\\
	Compute $MI\left(z^s ; l_{1: n}\right)$ through Eq. (\ref{eq:mi_zs});\\
	Make trajectory augmentation regarding $z_{1:n}^r$;\\
	Compute $MI\left(z_{1:n}^r ; l_{1: n}\right)$ through Eq. (\ref{eq:mi_zr});\\
	Update the parameters by maximizing Eq.(\ref{eq:obj});
	}
	}
\tcc{Task learning}
\Repeat{convergence}{
	\ForEach{$T$}
	{
	Compute each POI embedding via Eq.~(\ref{eq:emd});\\
	Obtain each hidden state $\boldsymbol{h}_\tau$ via Eq.(\ref{gru});\\
	Compute each $z_\tau^r$ via Eq.(\ref{eq:z_r});\\
	Compute $z^s$ based on the last hidden state;\\
	Model user historical trajectory via Eq.(\ref{eq:his});\\
	Obtain the predicted POI through Eq.(\ref{eq:pred});\\
    Update the parameters according to Eq.(\ref{eq:loss});
	}
    }
    \KwOut{Trained model.}
\end{algorithm}

\section{Experiments}
\label{Experiments}
We now conduct experiments to evaluate the performance of our proposed SSDL on four real-world datasets. 
\subsection{Experimental Settings}
\subsubsection{Datasets}
To facilitate reproducible results, we conduct all experiments on two publicly available LBS applications: Foursquare~\cite{yang2014modeling} and Gowalla~\cite{liu2014exploiting}. Foursquare contains check-ins in New York and Tokyo collected from 12 April 2012 to 16 February 2013. Each check-in has a timestamp, GPS coordinates and semantics about it. In Gowalla, we select the data from two cities, i.e., Los Angeles and Houston. Following previous studies~\cite{gao2019predicting,zhou2021self}, 
we filter out the POIs visited by fewer than eight times. For each user, we concatenate his/her all chronological check-ins and divide each trajectory into subsequence with the time interval of 24 hours. To specify whether check-ins are collected on weekdays or weekends, we further assign 48 time slots to each check-in time. We take each user's first 80\% trajectories as the training set, the remaining 20\% as the test set. The statistics of four datasets are summarized in Table~\ref{dataset-label}.

\renewcommand{\arraystretch}{0.8} 
\begin{table}[ht]
	\centering
    \setlength{\abovecaptionskip}{0pt}
	\setlength{\belowcaptionskip}{0pt}
	\caption{Statistics of the datasets.}
	\label{dataset-label}
\footnotesize
	\begin{tabular}{l|cccc}
		\toprule
		\textbf{City}& \textbf{Users} & \textbf{POIs}& \textbf{Check-ins} &\textbf{Trajectories}\\
		\hline
		\textbf{Tokyo}	&2102	&6789	&240056	&60365	\\ 
		\textbf{New York}	&990	&4211	&79006	&23252	\\ 
		\textbf{Los Angeles}	&2346	&8676	&195231	&61542	\\ 
		\textbf{Houston} &1351	&6994	&121502 &37514	\\ 
		\bottomrule
	\end{tabular}
\end{table}

\subsubsection{Baselines}
We compare our SSDL with several representative approaches for next POI prediction task.
\begin{itemize}[leftmargin=*,align=left]
	\item \textit{GRU}~\cite{cho2014learning} is a common approach for sequential data learning as its superiority in incorporating the semantics of long-term dependencies.
	
	\item \textit{ST-RNN}~\cite{liu2016predicting} is an RNN-based method that incorporates spatio-temporal contexts when predicting the next POI.
	
	\item \textit{HST-LSTM}~\cite{kong2018hst} employs sequence-to-sequence learning scheme to include spatial-temporal influence in LSTM and makes use of contextual information to enhance model performance for sparse data prediction.
	
	\item \textit{Flashback}~\cite{yang2021location} models sparse user mobility footprints by doing flashbacks on hidden states in RNNs. Especially, it explicitly employs the spatio-temporal contexts to search past hidden states with high predictive power. In our experiments, we take the GRU cell as the recurrent component in  Flashback for a fair comparison. 
	
	\item \textit{DeepMove}~\cite{feng2018deepmove} presents an attention-based RNN to encode human recent trajectories. Furthermore, it employs another RNN to learn user long-term preferences from historical trajectories.
	
	\item \textit{VANext}~\cite{gao2019predicting} proposes a novel variational attention mechanism to explore human periodic regularities. In addition, it employs a simple convolutional neural network rather than RNN to capture human long-term interests.
	
	\item \textit{PLSPL}~\cite{wu2020personalized} is a unified framework that jointly learns users' long- and short-term interests for next POI prediction.
	\item \textit{MobTCast}~\cite{xue2021mobtcast} is a Transformer-based approach that considers multiple semantic contexts behind check-ins to enhance the understanding of human mobility. Note that we remove the Social Context Extractor in MobCast as the social relationships are not available in our context.
	\item \textit{$\beta$-VAE}~\cite{burgess2018understanding} is a widely used representation learning method that is able to separate latent factors into different space by using an adjustable hyperparameter $\beta$ to the original VAE objective. In this study, we use GRU as the encoder and decoder network structure in $\beta$-VAE to model the temporal semantics.
	\item SML~\cite{zhou2021self} attempts to understand human mobility in a self-supervised learning manner. Especially, it leverages heuristic strategy to enumerate massive different views of original sparse trajectories for contrastive estimation.
\end{itemize}
\renewcommand{\arraystretch}{0.8}
\begin{table*}[t]
	\centering 
	\setlength{\abovecaptionskip}{0pt}
	\setlength{\belowcaptionskip}{0pt}
	\caption{Performance comparisons on four cities.
	}
	%\footnotesize
	%\fontsize{7.5}{9.}\selectfont 
	\scriptsize
	\label{tab:perform} 
	\setlength{\tabcolsep}{3.8mm}{\begin{tabular}{l|ccccc|ccccc}  
			\toprule
			\multicolumn{1}{l|}{\multirow{2}{*}{\textbf{Method}}}&  
			\multicolumn{5}{c|}{\textbf{Tokyo}}&\multicolumn{5}{c}{\textbf{New York}}\cr  
			\cline{2-6} \cline{7-11}  
			&ACC@1&ACC@5&ACC@10&AUC&MAP&ACC@1&ACC@5&ACC@10&AUC&MAP\cr  
			\hline
			
			GRU&13.11\	&27.88\	&34.28\	&88.01\ &7.36\     &15.37\ &31.73\ &36.10\ &81.40\ &8.59\ \cr
			
			ST-RNN&13.38\	&29.20\	&36.45\	&89.82\ &7.41\   &13.50\ &32.86\ &40.05\ &81.91\ &8.20\ \cr

			HST-LSTM&18.70\ &39.14\ &46.47\  &90.66\ &9.82\   &17.48\ &42.77\ &50.82\ &86.25\ &8.53\ \cr

			Flashback&18.23\ &39.42\ &46.66\ &90.40\ &10.47\   &22.22\ &49.52\ &57.11\ &87.74\ &13.59\ \cr
			
        	DeepMove&19.92\ &40.61\ &48.25\  &90.47\ &12.21\   &21.56\ &45.09\ &52.17\ &87.30\ &13.06\ \cr
        	
        	VANext&20.21\ &44.49\ &52.63\  &91.30\ &12.36\   &22.54\ &51.26\ &58.78\ &89.30\ &14.02\ \cr

            PLSPL  &20.19\	&43.64\	&52.45\	&91.37\ &\underline{12.86}\     &\underline{23.02}\ &53.33\  &63.34\  &89.21\ &\underline{14.83}\ \cr

           MobTCast  &19.58\	&43.41\	&51.95\	&89.95\ &11.67\     &22.37\ &\underline{54.31}\  &\underline{64.18}\  &88.59\ &14.03\ \cr

        	$\beta$-VAE &20.10\	&\underline{44.78}\	&\underline{53.89}\	&91.35\ &12.75\     &22.26\ &50.71\  &58.68\  &89.38\ &14.07\ \cr
        	
			SML&\underline{20.25}\ &{44.70}\ &{53.58}\  &\underline{91.48}\ &{12.51}\  &{22.62}\ &{52.16}\ &{60.18}\ &\underline{90.17}\ &{14.74}\ \cr
			
			\hline
			\textbf{SSDL}&\textbf{22.93} &\textbf{46.80} &\textbf{55.31} &\textbf{92.39} &\textbf{14.99} &\textbf{25.07}	&\textbf{56.78}	&\textbf{65.72}	&\textbf{90.72}	&\textbf{16.60}\cr
			\midrule
			\midrule
			\multicolumn{1}{l|}{\multirow{2}{*}{\textbf{Method}}}&  
			\multicolumn{5}{c|}{\textbf{Los Angeles}}&\multicolumn{5}{c}{\textbf{Houston}}\cr  
			\cline{2-6} \cline{7-11}  
			&ACC@1&ACC@5&ACC@10&AUC&MAP&ACC@1&ACC@5&ACC@10&AUC&MAP\cr  
			\hline			
			GRU&10.11\	&19.05\	&22.67\	&78.07\ &4.97\     &10.74\ &18.01\ &21.33\ &80.47\ &5.99\ \cr
			
			ST-RNN&10.01\	&19.30\	&23.64\	  &80.48\ &5.11\   &11.33\ &20.21\ &24.81\ &82.34\ &6.88\ \cr

			HST-LSTM&12.01\ &23.97\ &29.04\  &82.57\ &5.29\   &13.41\ &22.86\ &27.21\ &82.58\ &6.58\ \cr

			Flashback&13.81\ &25.60\ &30.32\ &83.74\ &7.65\   &14.37\ &24.52\ &28.70\ &84.54\ &8.79\ \cr
			
        	DeepMove&13.31\ &25.73\ &30.35\  &82.41\ &7.26\   &14.13\ &24.59\ &29.03\ &83.29\ &8.46\ \cr
        	
        	VANext&14.36\ &27.91\ &33.16\  &86.22\ &7.73\   &14.88\ &26.78\ &31.44\ &86.06\ &8.31\ \cr
        	
        	PLSPL  &\underline{14.92}\	&28.26\	&\underline{33.86}\	&84.34\ &\underline{7.97}\     &\underline{16.06}\ &\underline{29.22}\  &\underline{34.67}\  &86.11\ &\underline{9.74}\ \cr

           MobTCast  &14.30\	&\underline{28.62}\	&33.38\	&83.41\ &7.65\     &15.40\ &28.27\  &32.87\  &83.87\ &8.66\ \cr
        	
        	$\beta$-VAE &14.39\	&27.43\	&32.96\	&85.95\ &7.72\     &15.00\ &27.09\  &31.89\  &86.10\ &8.23\ \cr
        	
			SML&{14.77}\ &{28.12}\ &{33.35}\  &\underline{86.38}\ &{7.86}\  &{15.15}\ &{27.46}\ &{32.31}\ &\underline{86.48}\ &{9.17}\ \cr

			\hline
			\textbf{SSDL}&\textbf{15.94} &\textbf{31.02} &\textbf{36.80} &\textbf{86.98} &\textbf{8.72} &\textbf{16.91}	&\textbf{30.92}	&\textbf{36.56}	&\textbf{87.30}	&\textbf{10.70}\cr
			\bottomrule 
			
		\end{tabular}
		}
\end{table*}

\subsubsection{Metrics}
To evaluate the performance of our proposed SSDL, we follow most of the previous studies~\cite{gao2019predicting,zhou2021self,wu2020personalized} and select three commonly used metrics to compare with the baselines. We first use the ACC@$K$ to evaluate the recommendation performance. In this paper, we report the different testing results of $K={1,5,10}$. Additionally, we report area under the ROC curve (AUC) and mean average precision (MAP) metrics that are frequently used in classification tasks.
\subsubsection{Implementation Details}
We implement our SSDL and baselines in Python. All methods are based on the Torch library and accelerated by one NVIDIA GTX 1080 GPU. Besides, we choose Adam~\cite{kingma2014adam} to train all deep learning methods. In disentanglement learning, the learning rate is initialized as 0.01. We set the coefficient $\alpha$ of KL terms to 1. Besides, $\beta$ and $\gamma$ are fixed to be 1 and 0.1. In task learning, the learning rate is initialized with 5e-4. The dropout rate is set as 0.5, and the batch size is 32. The hidden size of the self-attention network is set to 300. In addition, we set dimension of $z^s$ to 256, while $z^r_{1:n}$ to 32.

\subsection{Performance Comparisons}
Table~\ref{tab:perform} reports the performance of different approaches on the datasets of four cities, where the best achievement is highlighted with \textbf{bold} and the second best is marked with \underline{underline}. Specifically, we have the following observations.

We can find that ST-RNN does not provide us with competitive achievements compared to GRU although it considers the spatial and temporal constraints. The plausible reason is that the sparsity issue of check-in data heavily affects the distillation of semantics contexts such as geographical distance. Meanwhile, relying on simple spatio-temporal features and regarding the next POI as the solo supervision usually results in an inference bias or uncertainty problem due to the boundary of available training datasets. To mitigate the data sparsity issue, HST-LSTM which combines spatial and temporal factors with a gate mechanism is able to boost the capture of human mobility patterns by a large margin. Furthermore, HST-LSTM models the periodicity of consecutive check-ins in an end-to-end manner, which brings an encouraging prospect for us to learn the complex distribution behind historical trajectories. Compared with HST-LSTM, which directly adds spatio-temporal factors to hidden states, Flashback achieves competitive performance because it explicitly uses a rich spatio-temporal context to search for past hidden states with high predictive power to predict the next POI.

As for DeepMove and VANext, they both attempt to correlate a certain user's recent trajectory and historical trajectory to accurately discover individual periodicity. Our experiments show that they achieve higher gains than models (e.g., ST-RNN) that only consider the past few check-ins. Furthermore, VANext, the first variational inference approach to model human trajectories using a prior assumption, outperforms DeepMove due to the relief of the inherent uncertainty of user mobility. The paradigm of PLSPL is similar to DeepMove, but it operates an attention mechanism to evaluate the importance of each POI in a user's historical check-ins, aiming at exploring the tastes of different users. We can find that PLSPL performs better than DeepMove. In addition, MobTCast is a Transformer-based approach that uses self-attention to study the interactive signals between POIs in a given trajectory, as well as multiple semantic contexts, such as category and temporal semantics. We obtain similar performance results compared to PLSPL, indicating that considering multiple semantic contexts does help to discover users' future check-in intentions.

$\beta$-VAE is a popular disentanglement learning method that also obtains promising results, which suggests that employing the latent variables produced by variational Bayesian does help in understanding the inherent generative factors underlying human mobility. As for SML, it is the first self-supervised learning solution for the next POI prediction, achieving the best gains on AUC among the baselines. The reason is that it primarily seeks to produce massive synthetic trajectories for data augmentation and leverage contrastive learning to study the diversity of human moving intents behind existing historical check-ins.

In general, our proposed SSDL significantly outperforms the compared approaches by a relatively large margin across the four cities. For instance, SSDL respectively yields 8.57\% and 11.94\% averaged improvement over the best baseline regarding ACC@1 and MAP. This observation demonstrates the superiority of the self-supervised disentanglement learning paradigm in our SSDL. 
\begin{figure*}[ht]
    \centering
    \includegraphics[width=0.255\textwidth]{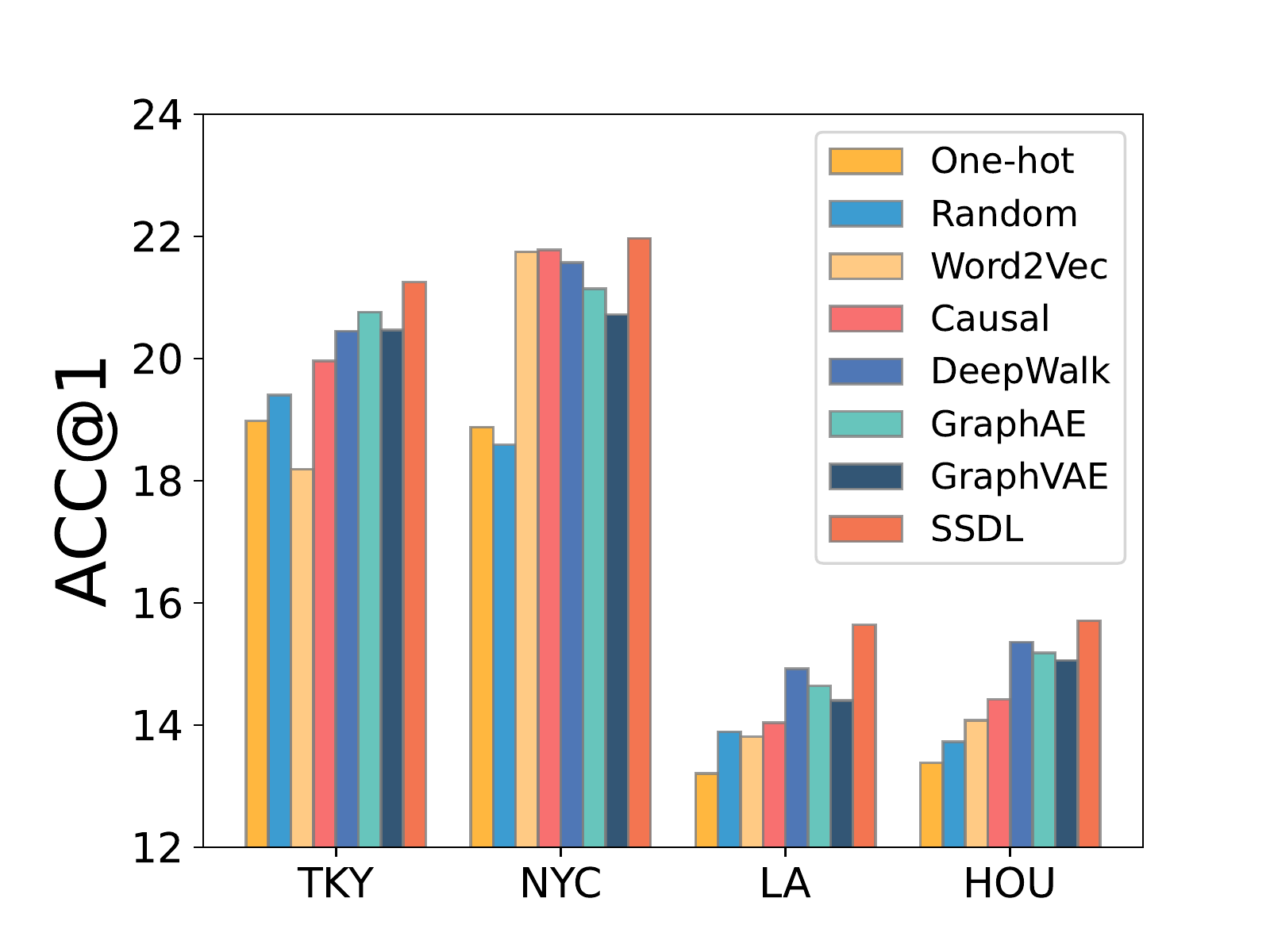}
    \hspace{-0.4cm}
    \includegraphics[width=0.255\textwidth]{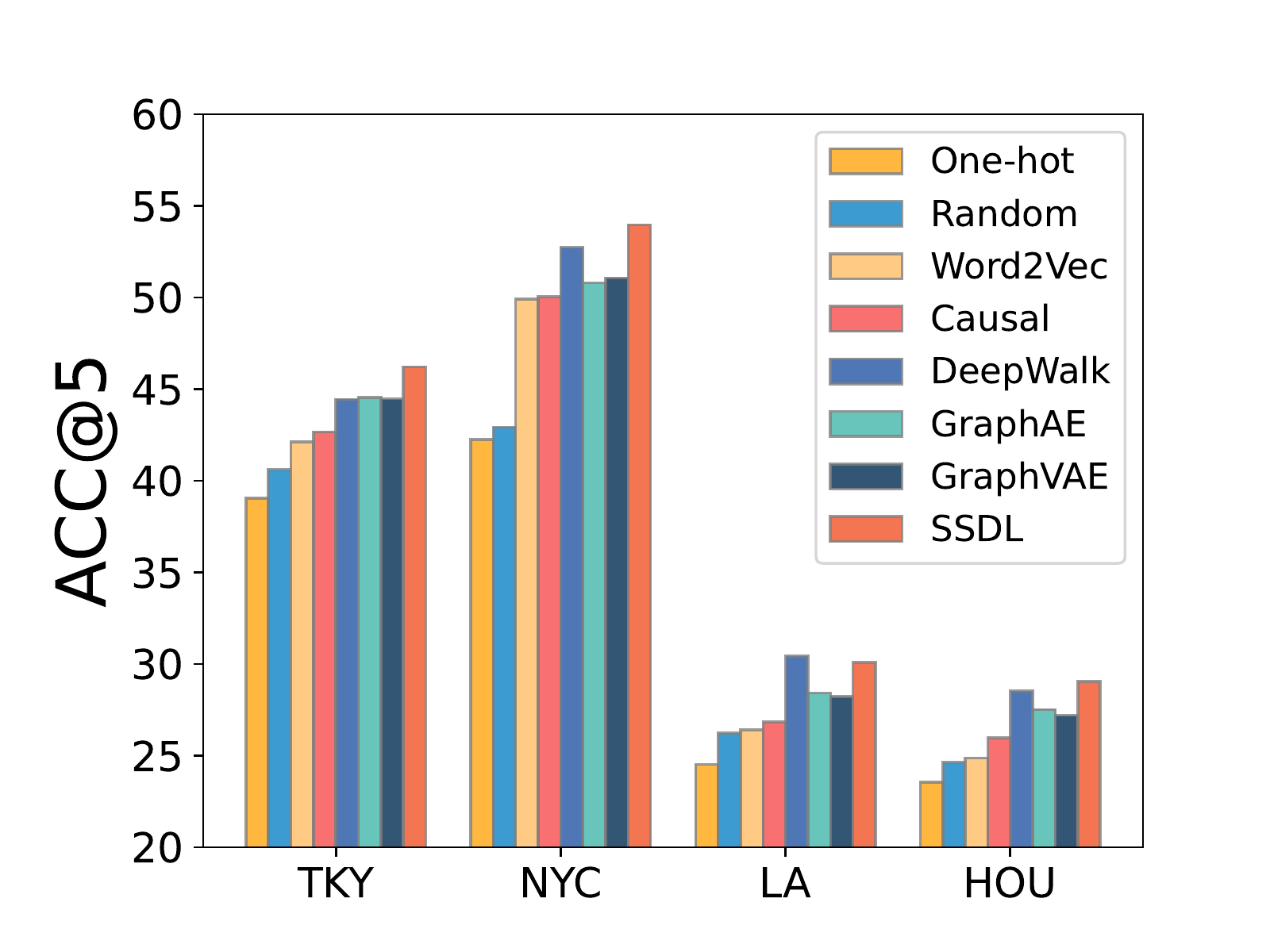}
     \hspace{-0.4cm}
    \includegraphics[width=0.255\textwidth]{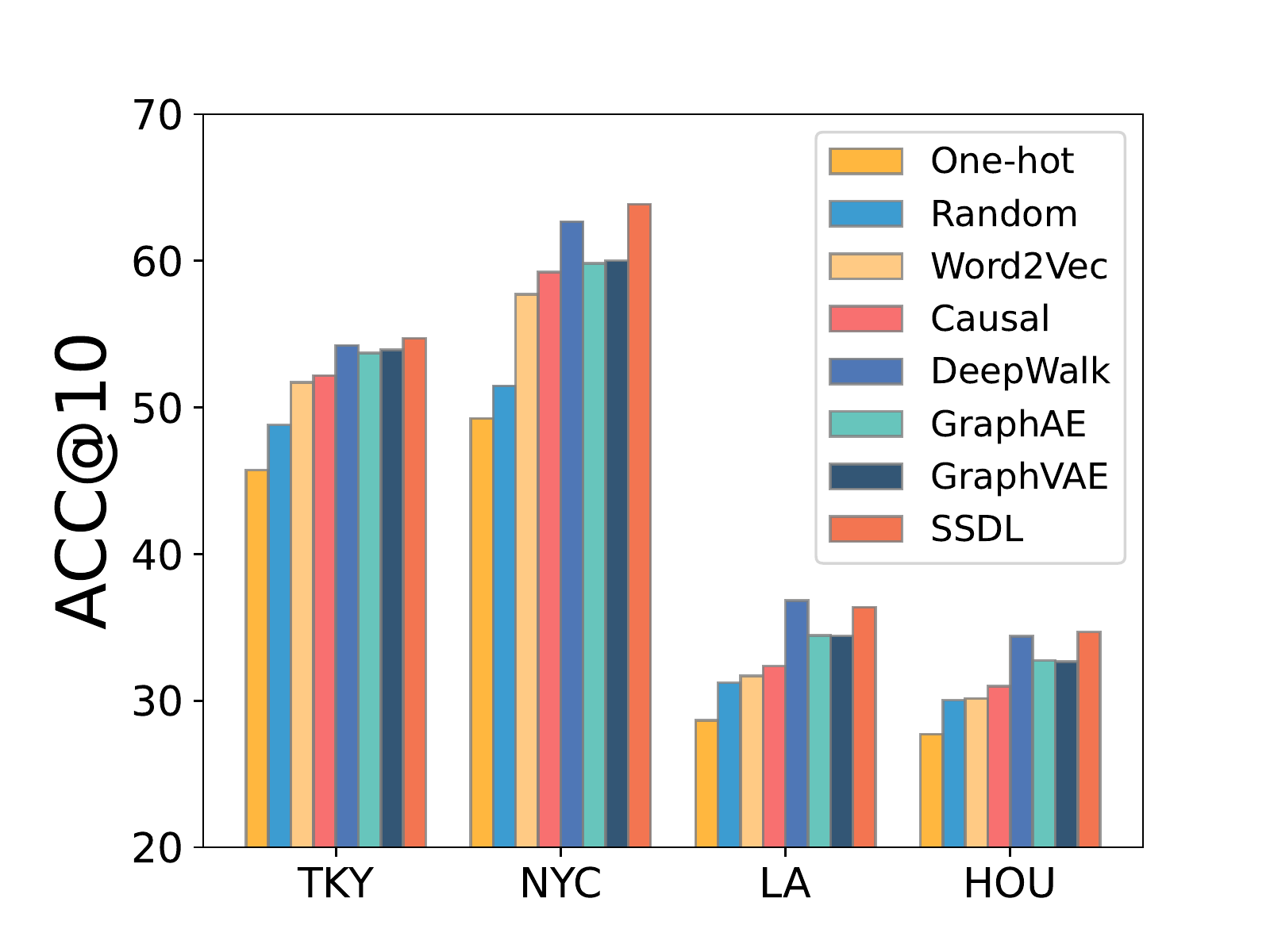}
     \hspace{-0.4cm}
    \includegraphics[width=0.255\textwidth]{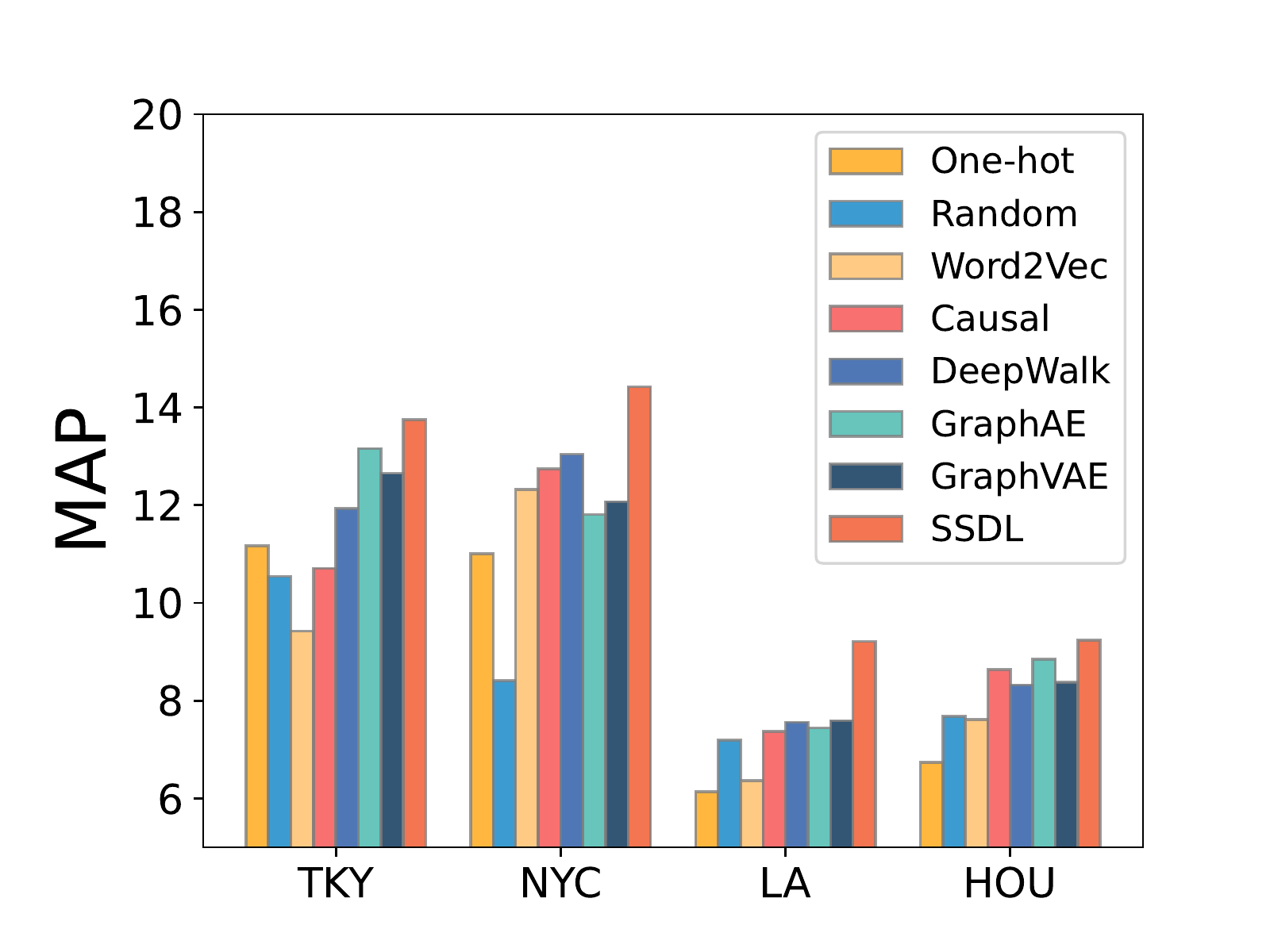}
    \caption{Effect of POI embedding.}
    \label{fig:poi-em}
\end{figure*}
\begin{figure*}[ht]
%\vspace{-0.2cm}
    \centering
    \includegraphics[width=0.255\textwidth]{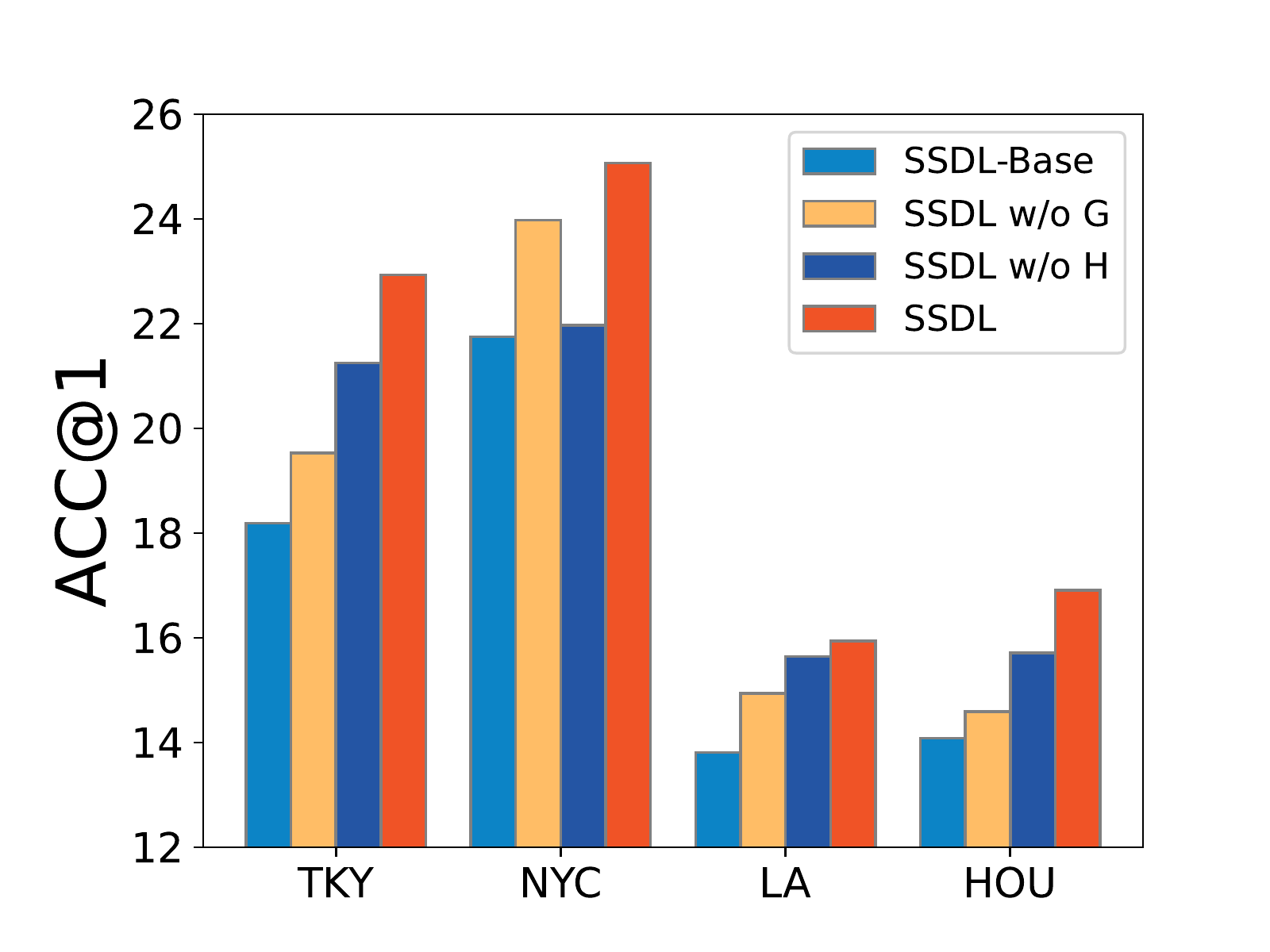}
    \hspace{-0.4cm}
    \includegraphics[width=0.255\textwidth]{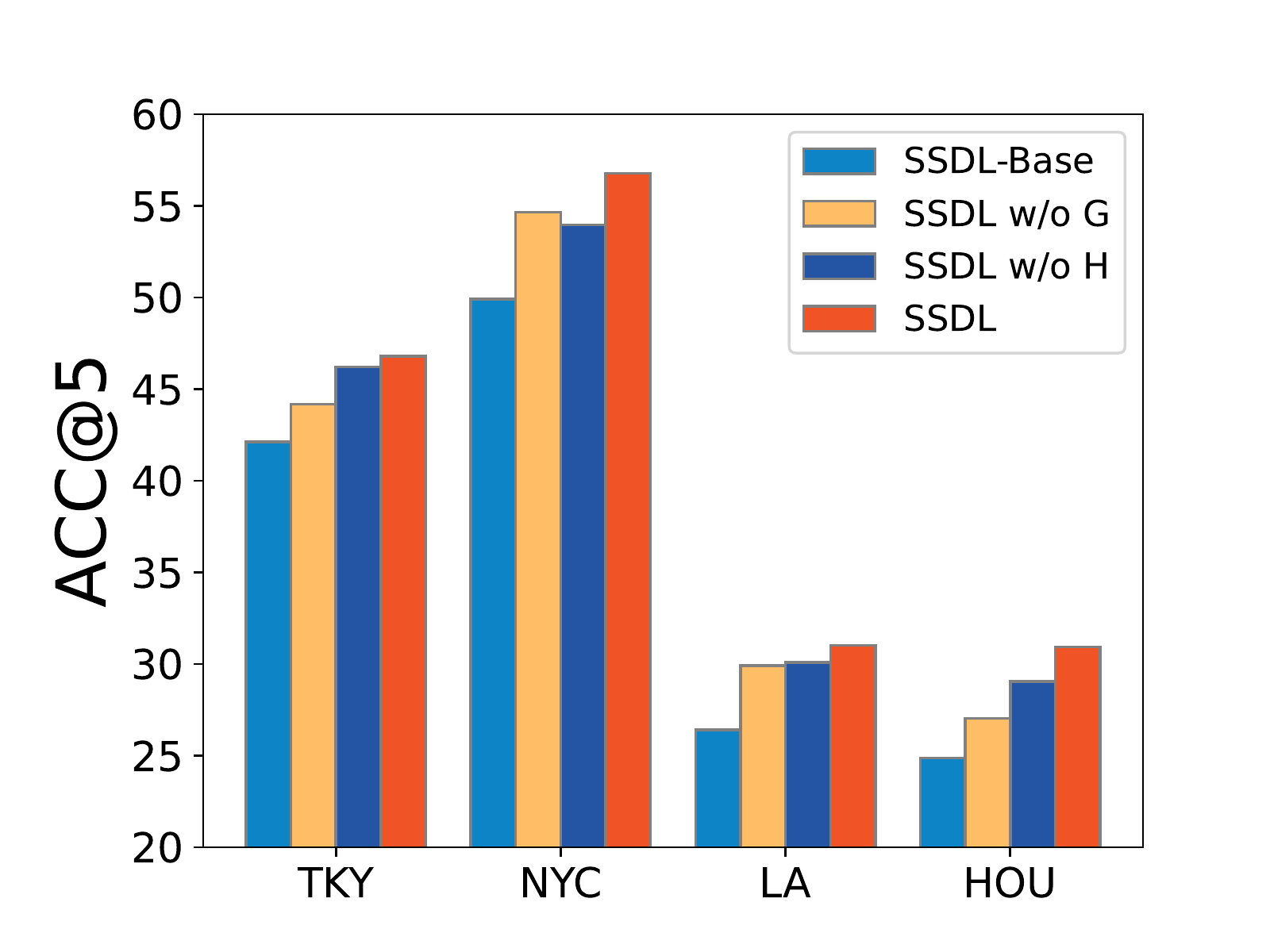}
     \hspace{-0.4cm}
    \includegraphics[width=0.255\textwidth]{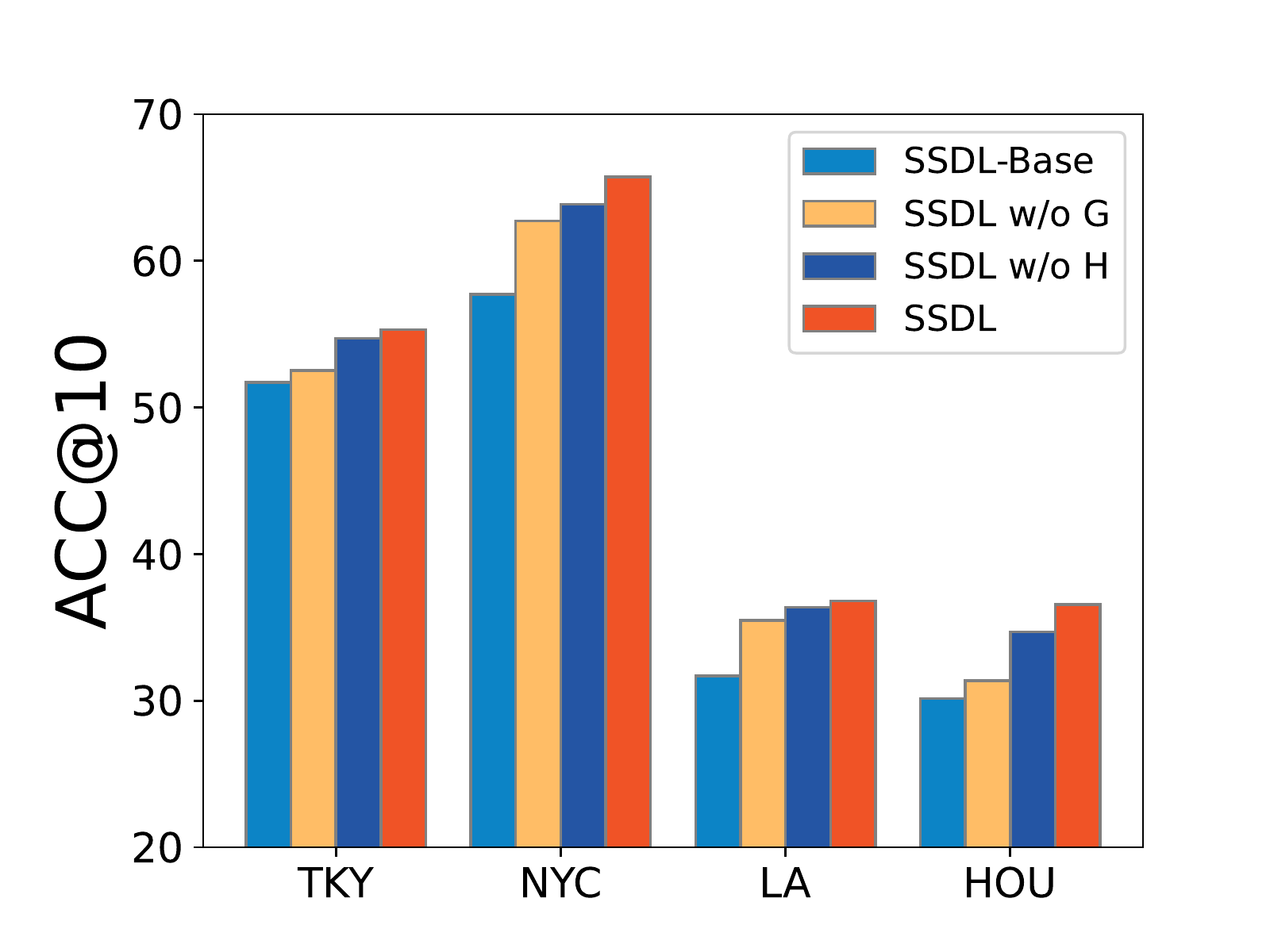}
     \hspace{-0.4cm}
    \includegraphics[width=0.255\textwidth]{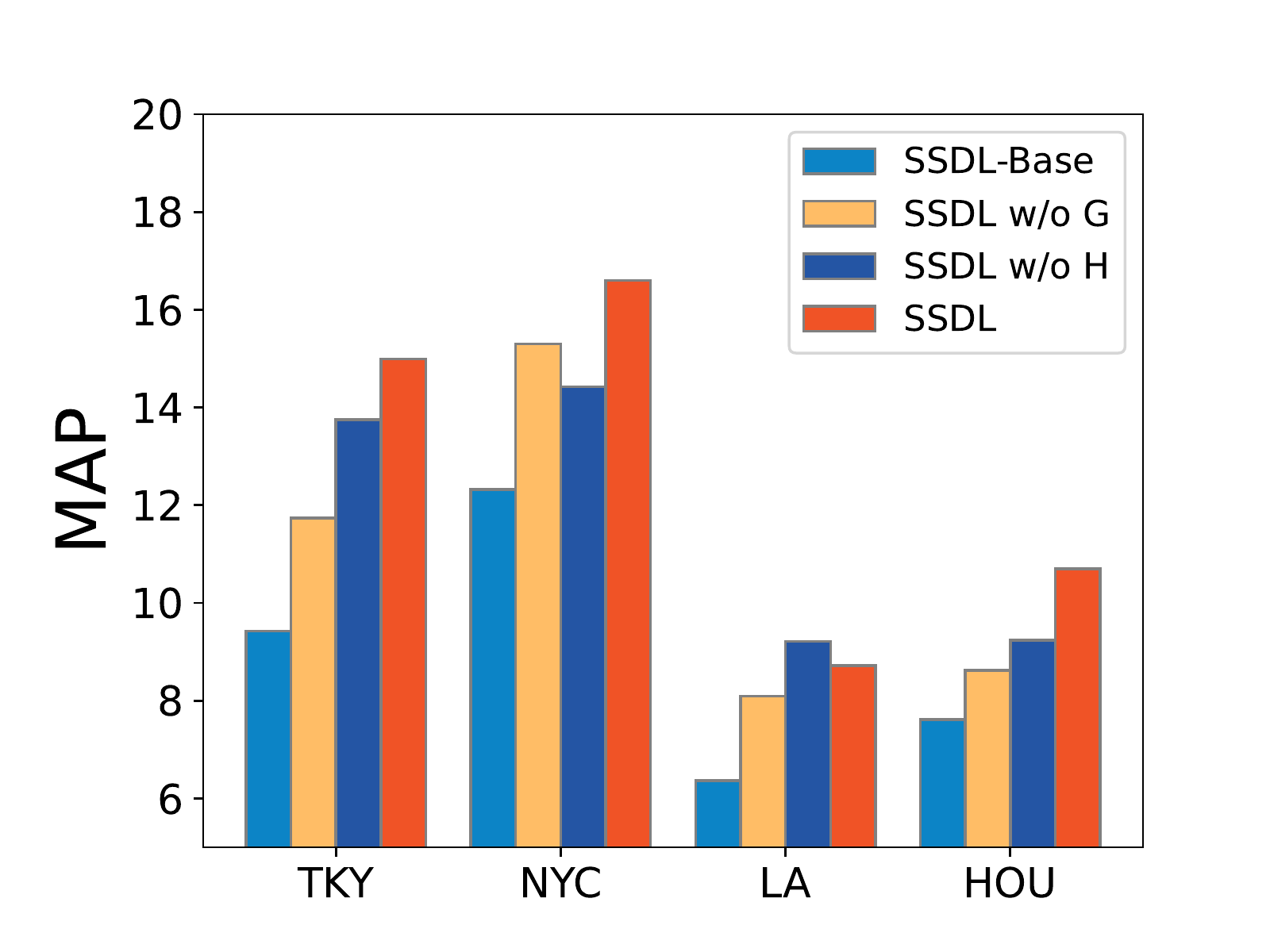}
    \caption{Effects of different components in SSDL.}
    \label{fg:module}
\end{figure*}
\subsection{Ablation Study}
To evaluate the contribution of different components of SSDL, we devise several variants regarding SSDL from two aspects, i.e., POI embedding and trajectory embedding. First, we select 7 popular embedding methods to scrutinize the efficacy of our POI embedding. 
\begin{itemize}[leftmargin=*,align=left]
    \item \textbf{One-Hot}~\cite{feng2018deepmove} is the simplest method that maps each POI to a unique vector without any semantic information. 
    \item \textbf{Random} uses a dense matrix sampled from a Gaussian distribution to represent the POIs.
    \item \textbf{Word2vec} is a popular embedding technique in NLP, aiming at exploring the surrounding context of a given word. Also, it has successfully applied in POI embedding~\cite{gao2017identifying,feng2017poi2vec}. We implement the skip-gram model for POI embedding.
    \item \textbf{Causal}~\cite{gao2019predicting} is a variant of word2vec that treats the previous footprints of the current POI as its semantic context to incorporate human practical transitional behaviors.
    \item \textbf{Deepwalk}~\cite{gao2022contextual}is a data augmentation method that builds a POI graph to integrate users' historical visiting interests and geographical proximity and then leverages the skip-gram technique for POI embedding.
    \item \textbf{GraphAE} is popular method for node embedding. In this paper, we treat each POI as a node and build the same graph as Deepwalk for POI embedding.
    \item \textbf{GraphVAE} is a variant of GraphAE, taking the advantage of VAE for POI embedding.
\end{itemize}
Fig.~\ref{fig:poi-em} reports the performance of SSDL using different POI embedding methods. We can observe that our embedding method achieves the best results on the vast majority of metrics across the four cities, which indicates its higher effectiveness in capturing multiple human interests behind historical trajectory data.

Next, we turn to investigate the effectiveness of devised components in SSDL. Herein, we conduct the experiments with three SSDL variants. The details are shown as follows:
\begin{itemize}[leftmargin=*,align=left]
    \item \textbf{SSDL-Base} is a basic model that removes both graph-based embedding and mutual information regularization of SSDL. Instead, we use the word2vec technique for POI embedding.
    \item \textbf{SSDL w/o G} only removes the graph-based embedding and use the word2vec for POI embedding.
    \item \textbf{SSDL w/o H} only removes the mutual information regularization of SSDL.
\end{itemize}
Fig.~\ref{fg:module} illustrates the performance of variants on four cities. First, we can find that removing any modules would bring significant performance degradation, suggesting that both modules in our SSDL benefit to enhance POI prediction. Second, SSDL-Base performs worse than SSDL w/o H across all cities, demonstrating that considering multiple common interests behind historical check-in data are useful to discover human mobility patterns. 
Third, SSDL w/o G outperforming SSDL-Base proves that our self-supervised disentanglement learning is an effective module to provide promising representations for task inference. %Finally,

\subsection{Disentanglement Interpretability}
In this part, we focus on studying the disentangled representations from the interpretability aspect. We first investigate whether $z^s$ and $z_{1:n}^r$ can be well extracted from original trajectories and reflect human time-invariant periodicity/habits and time-varying interests, respectively. 
To this end, we randomly sample eight different users' trajectories and change their orders to generate several groups of trajectories. Then, we use the TSNE toolkit~\cite{van2008visualizing} to visualize the distribution time-invariant representations. We can find that the representations of $z^s$ produced by SSDL are grouped well, demonstrating that it can successfully separate the time-invariant factors to uncover the inherent preference of users that are not influenced by temporal factors. For $z_{1:n}^r$, we visualize the distribution of the last states $z_n^r$ for simplicity, we can find they are entangled, indicating that they are really affected by the temporal factors. Therefore, we conclude that $z^s$ and $z_{1:n}^r$ indeed play well the roles of time-invariant and time-varying representations, respectively.
\begin{figure}[ht]
\vspace{-0.4cm}
\centering
\subfigure[$z^s$ in SSDL.]{
    \includegraphics[width=0.235\textwidth]
    {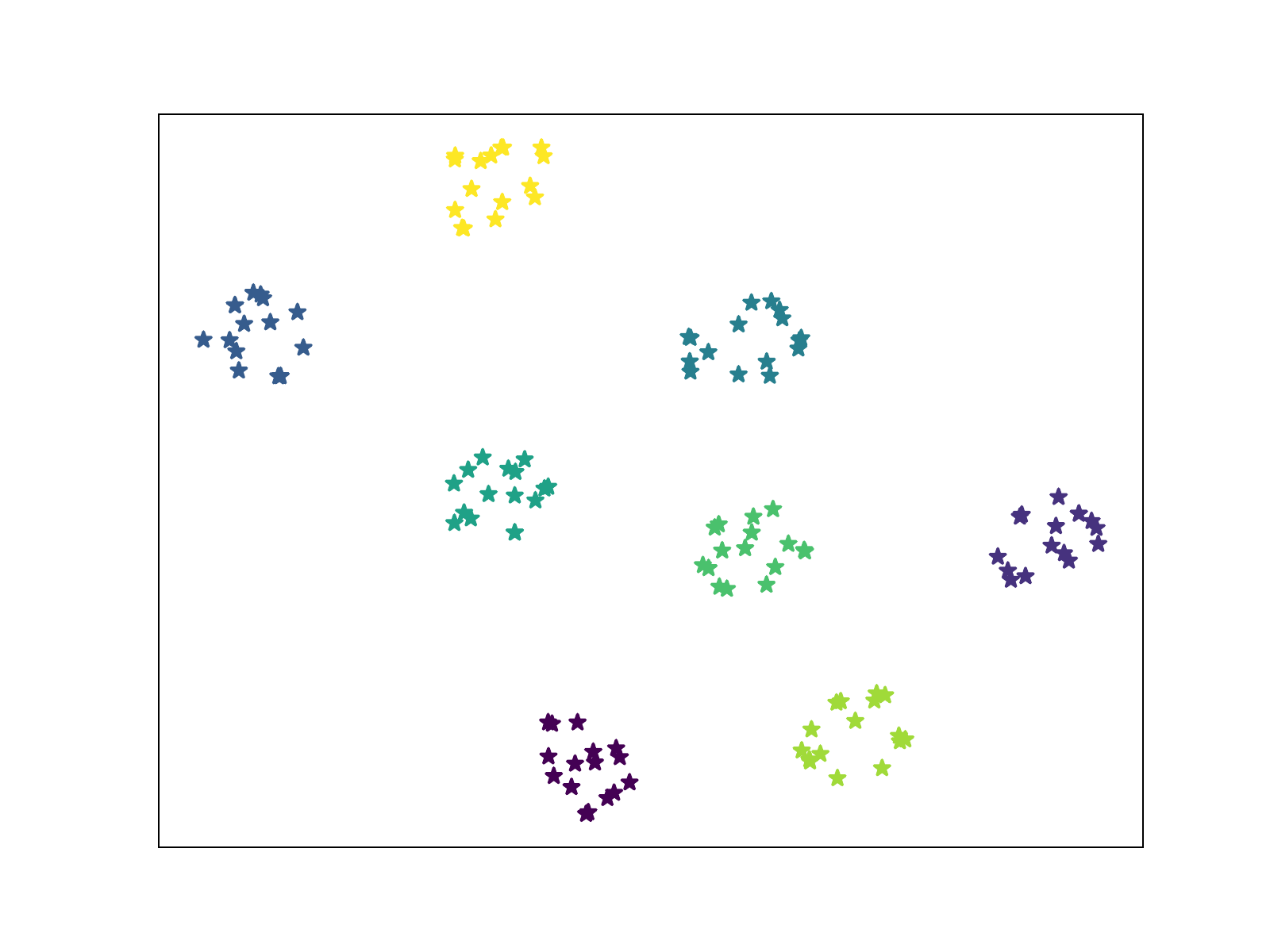}
}
\hspace{-0.55cm}
\subfigure[$z_n^r$ in SSDL.]{
    \includegraphics[width=0.235\textwidth]
    {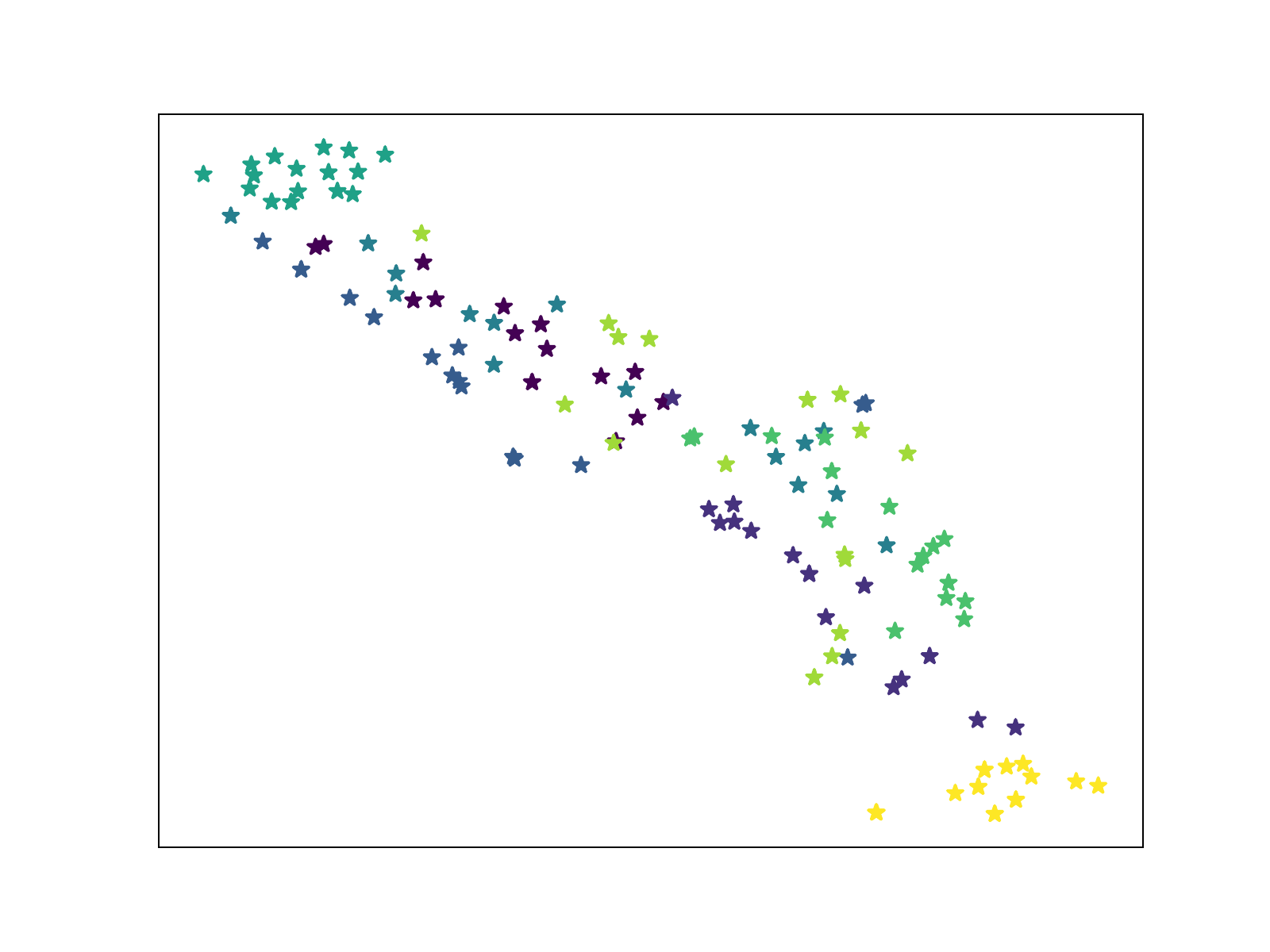}
}
    \caption{The visualization of latent representations.}
    \label{fg:vi-after}
    %\vspace{-0.2cm}
\end{figure}

\begin{figure}[ht]
\vspace{-0.4cm}
\centering
\subfigure[$\beta$-VAE.]{
    \includegraphics[width=0.235\textwidth]
    {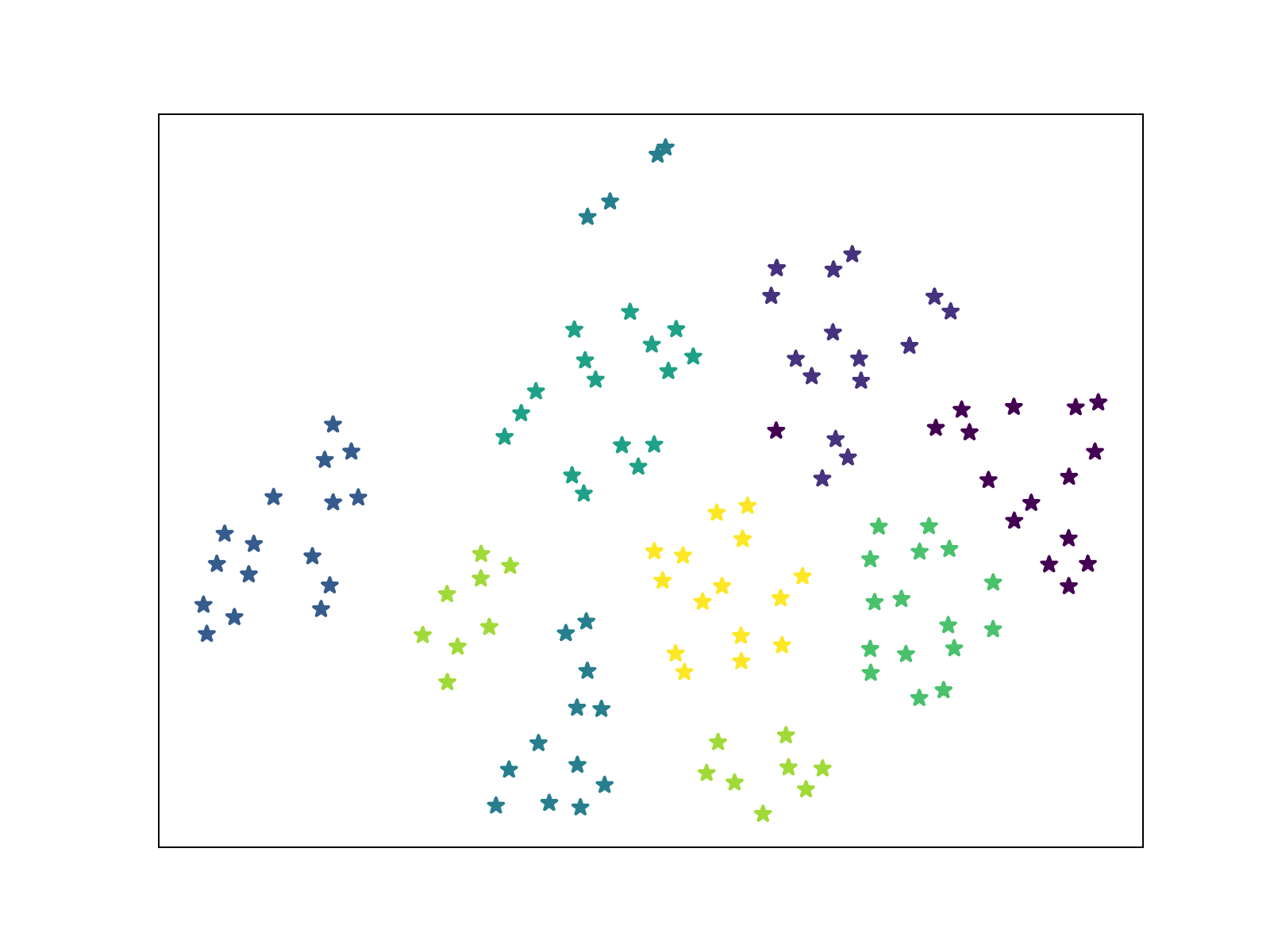}
    \label{fg:vi-aug-vae}
}
\hspace{-0.55cm}
\subfigure[SSDL w/o Aug.]{
    \includegraphics[width=0.235\textwidth]
    {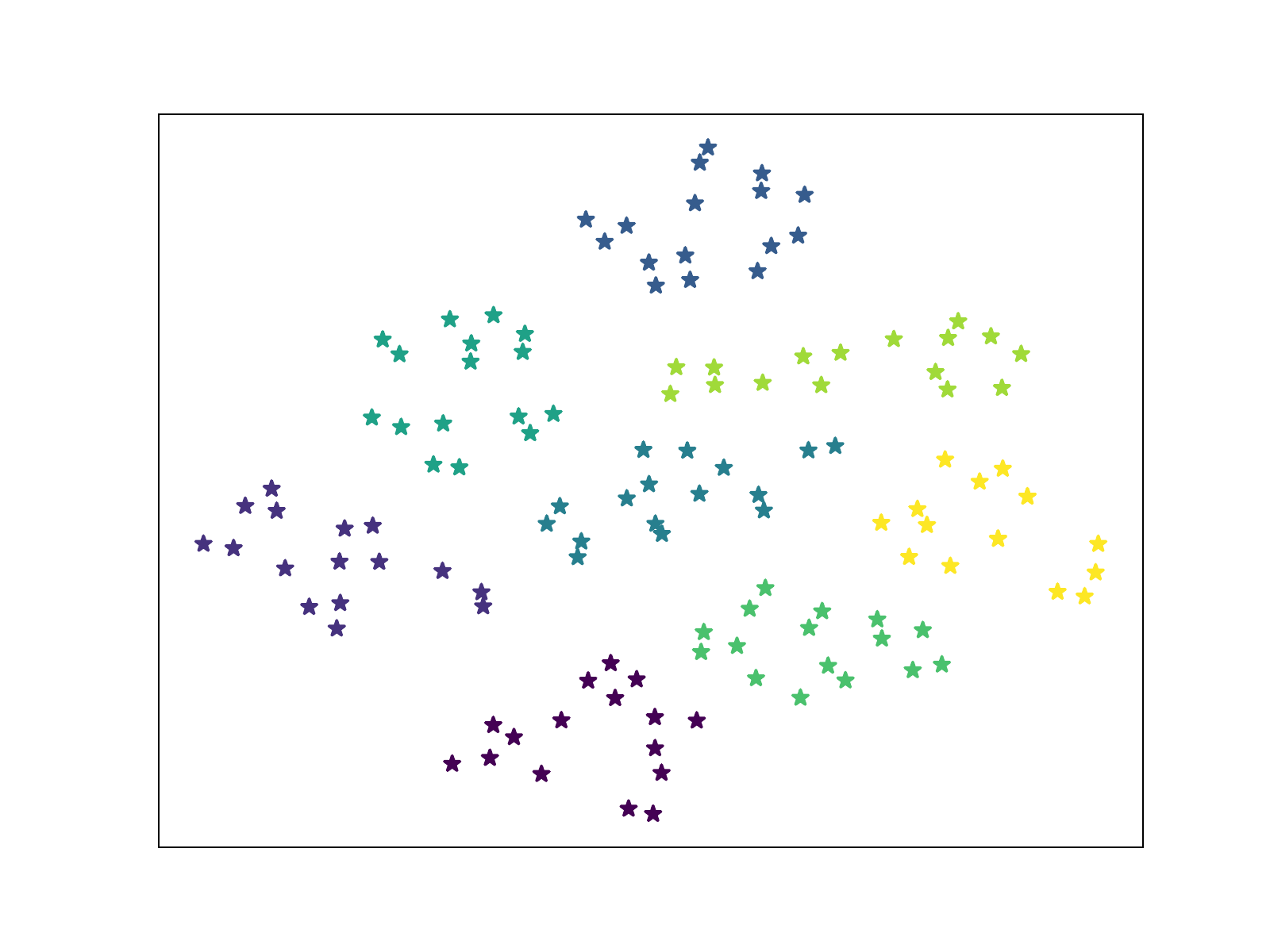}
    \label{fg:vi-aug-none}
}
\hspace{-0.55cm}
\subfigure[SSDL w/o Random.]{
    \includegraphics[width=0.235\textwidth]
    {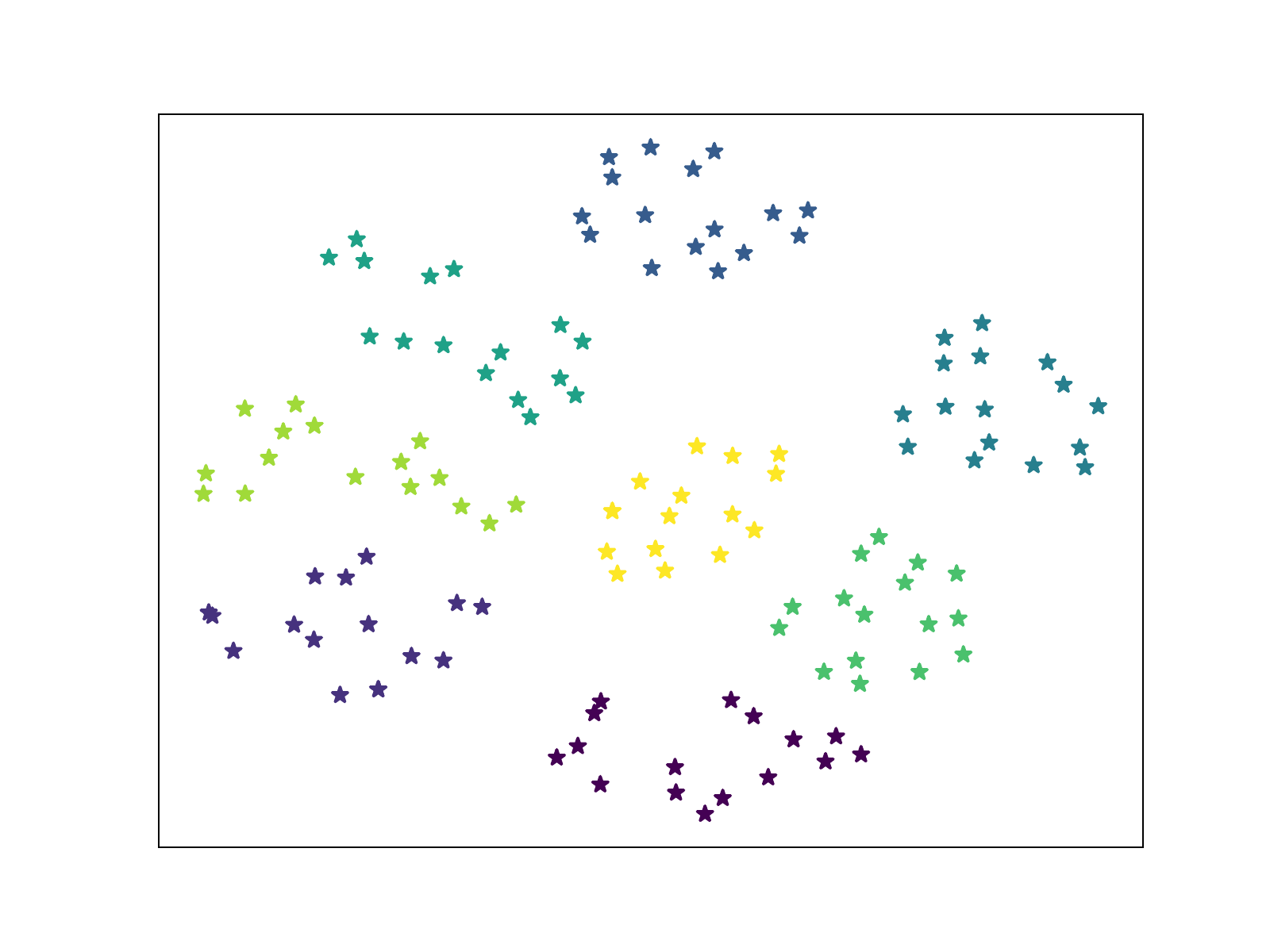}
    \label{fg:vi-aug-rand}
}
\hspace{-0.55cm}
\subfigure[SSDL.]{
    \includegraphics[width=0.235\textwidth]
    {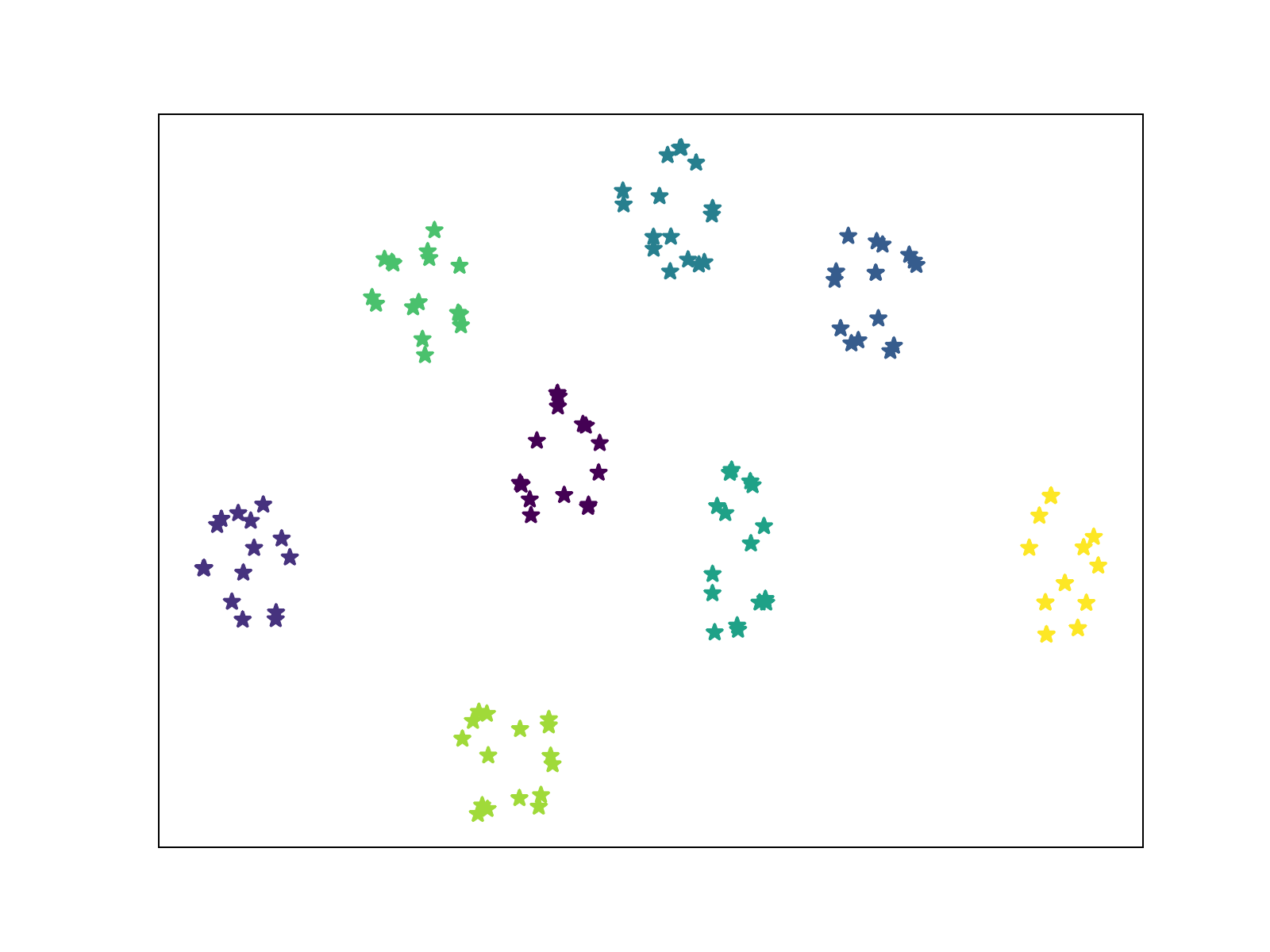}
    \label{fg:vi-aug-ssdl}
}
    \caption{The impact of augmentations on $z^s$.}
    \label{fg:vi-aug}
\end{figure}

Besides, we also study the impact of our data augmentation approaches from a visualization perspective. We visualize the $z^s$ distribution of randomly sampled trajectories of eight different users after task training. As shown in Fig.~\ref{fg:vi-aug-vae}, we can find $\beta$-VAE can only separate the representations with a small margin. Fig.~\ref{fg:vi-aug-none} presents the  results of SSDL without any data augmentations, and Fig.~\ref{fg:vi-aug-rand} shows the results of SSDL that has no augmentation for time-invariant factors. Compare to Fig.~\ref{fg:vi-aug-ssdl}, we can clearly find that both augmentations used in SSDL can significantly help us distinguish the trajectory representations of different users. This observation further suggests that different user movement patterns can be well refined by our SSDL.

\subsection{Sensitivity Analysis}
\begin{figure}[ht]
%\vspace{-0.4cm}
\centering
\includegraphics[width=0.25\textwidth]
{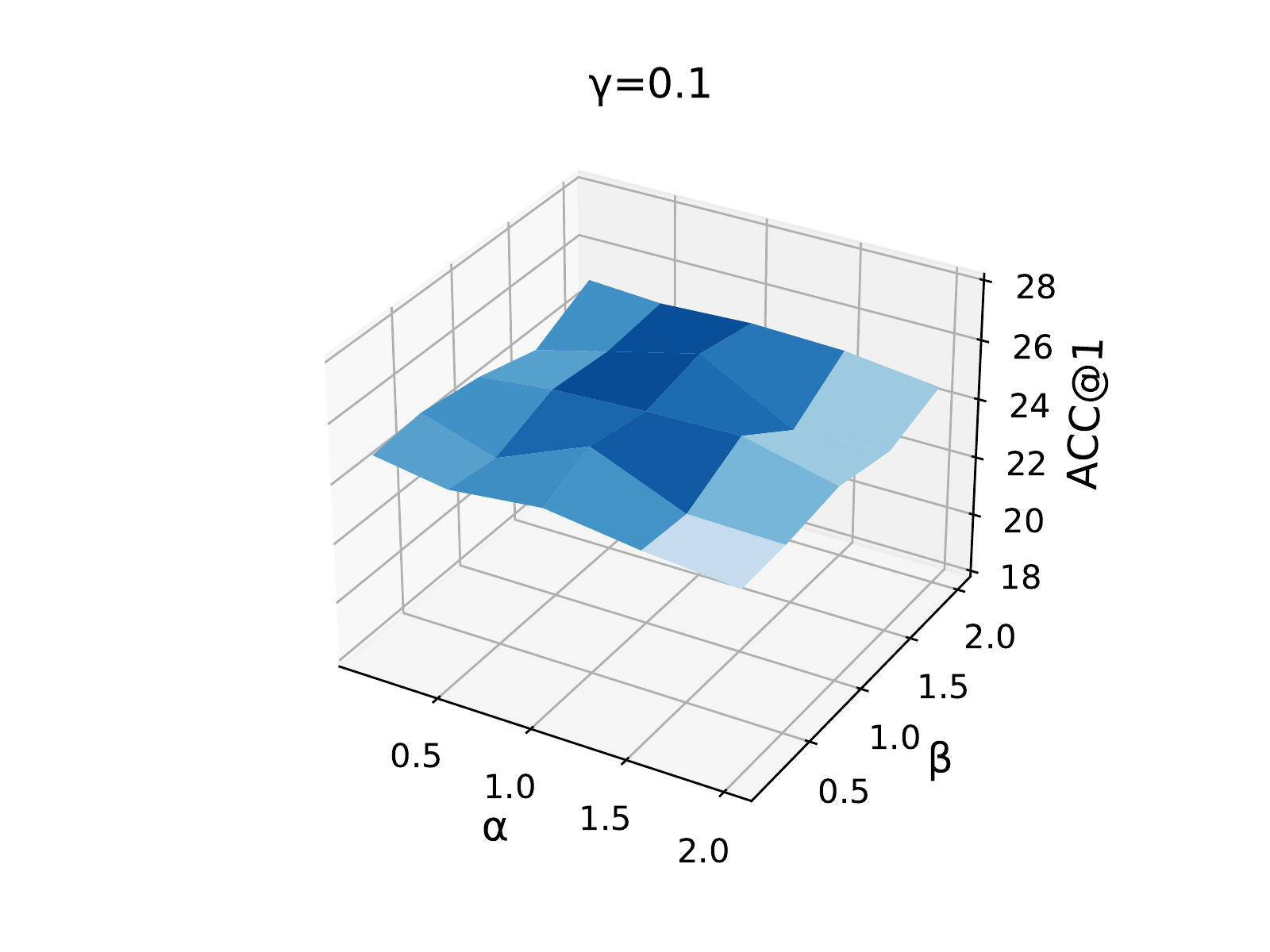}
\hspace{-0.45cm}
\includegraphics[width=0.25\textwidth]
{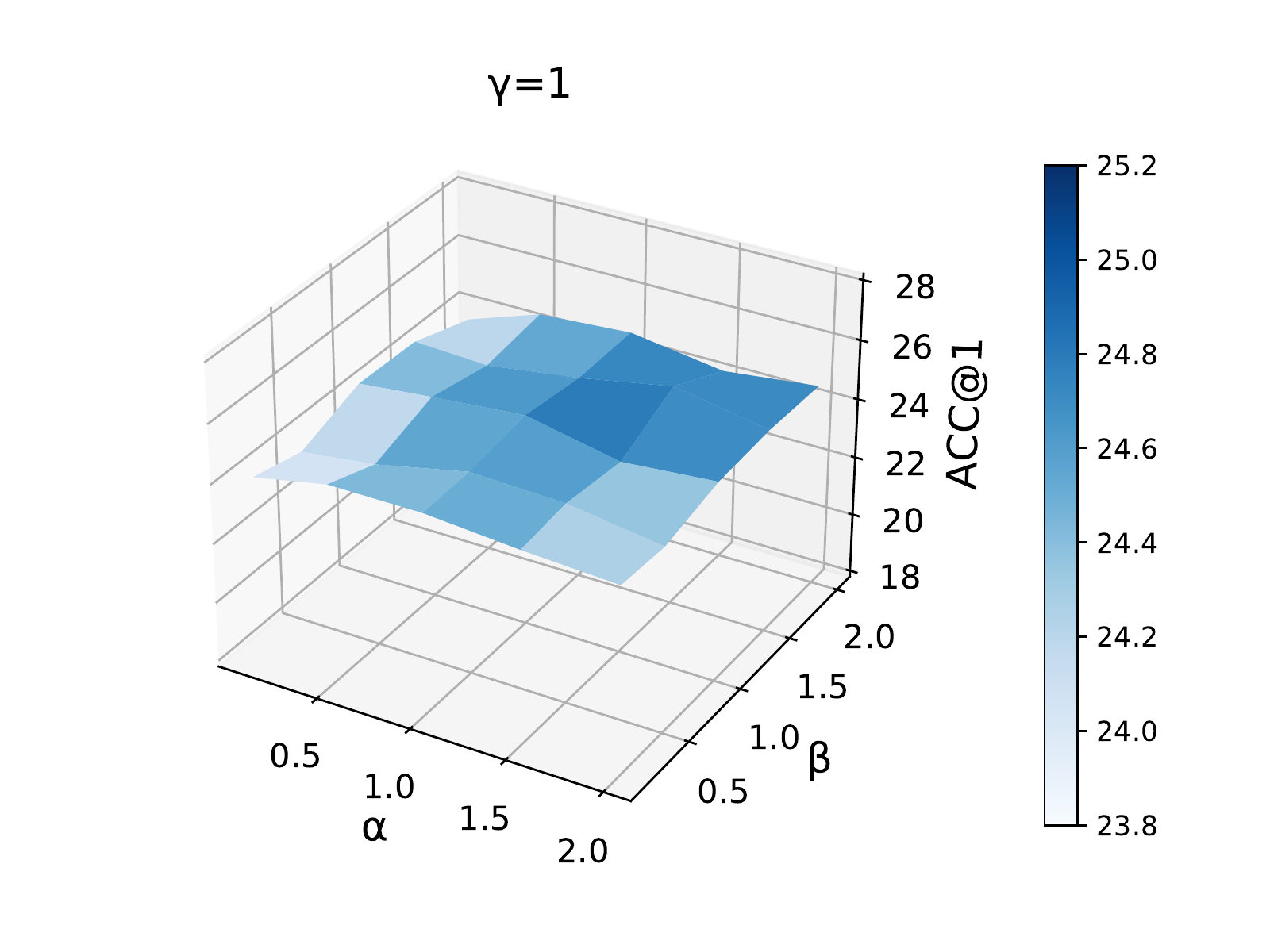}
\caption{The influence of weight coefficients in New York dataset.}
    \label{fg:weight_coefficients}
\end{figure}

\begin{figure}[ht]
\centering
\includegraphics[width=0.17\textwidth]
{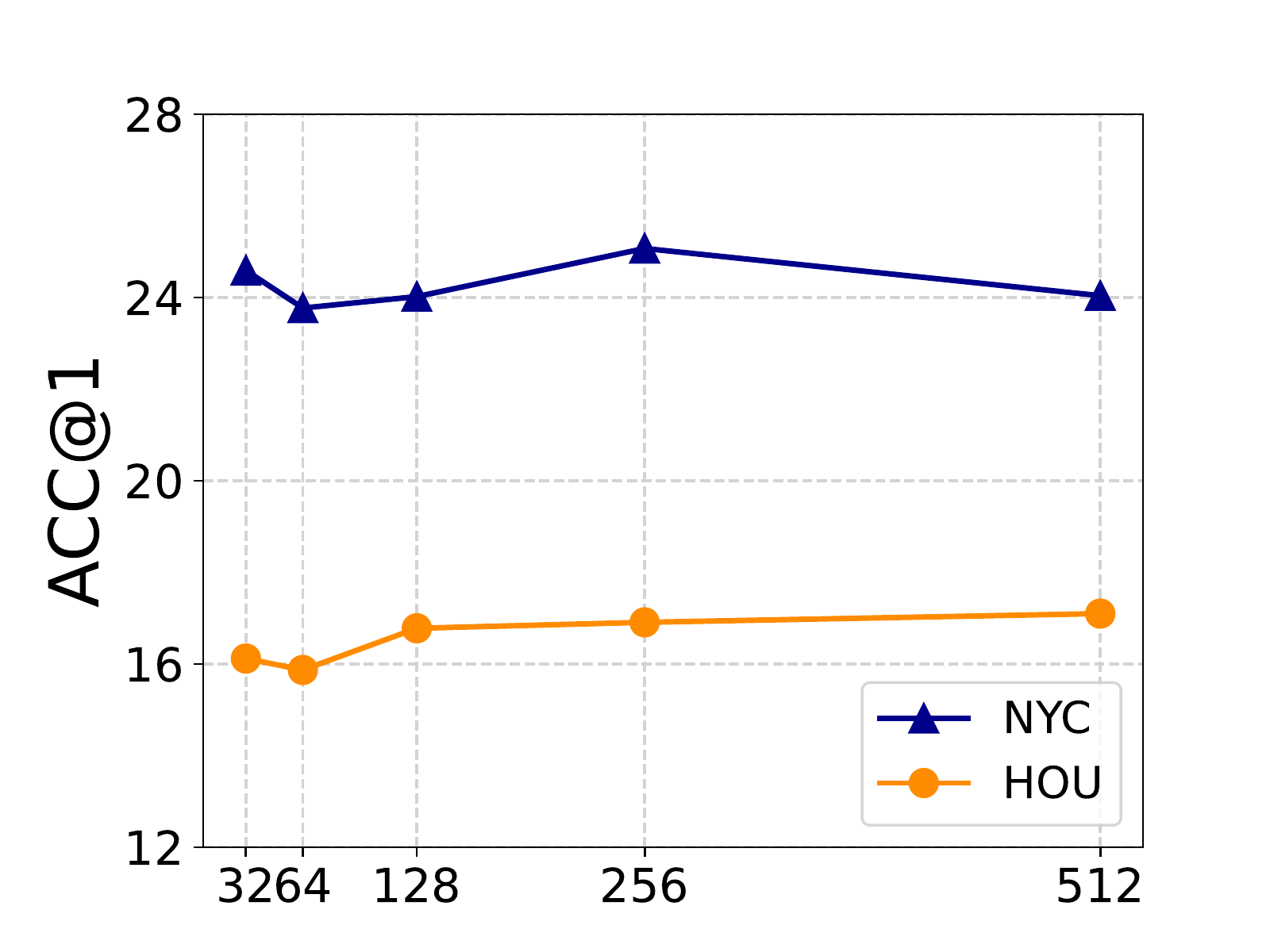}
\hspace{-0.45cm}
\includegraphics[width=0.17\textwidth]
{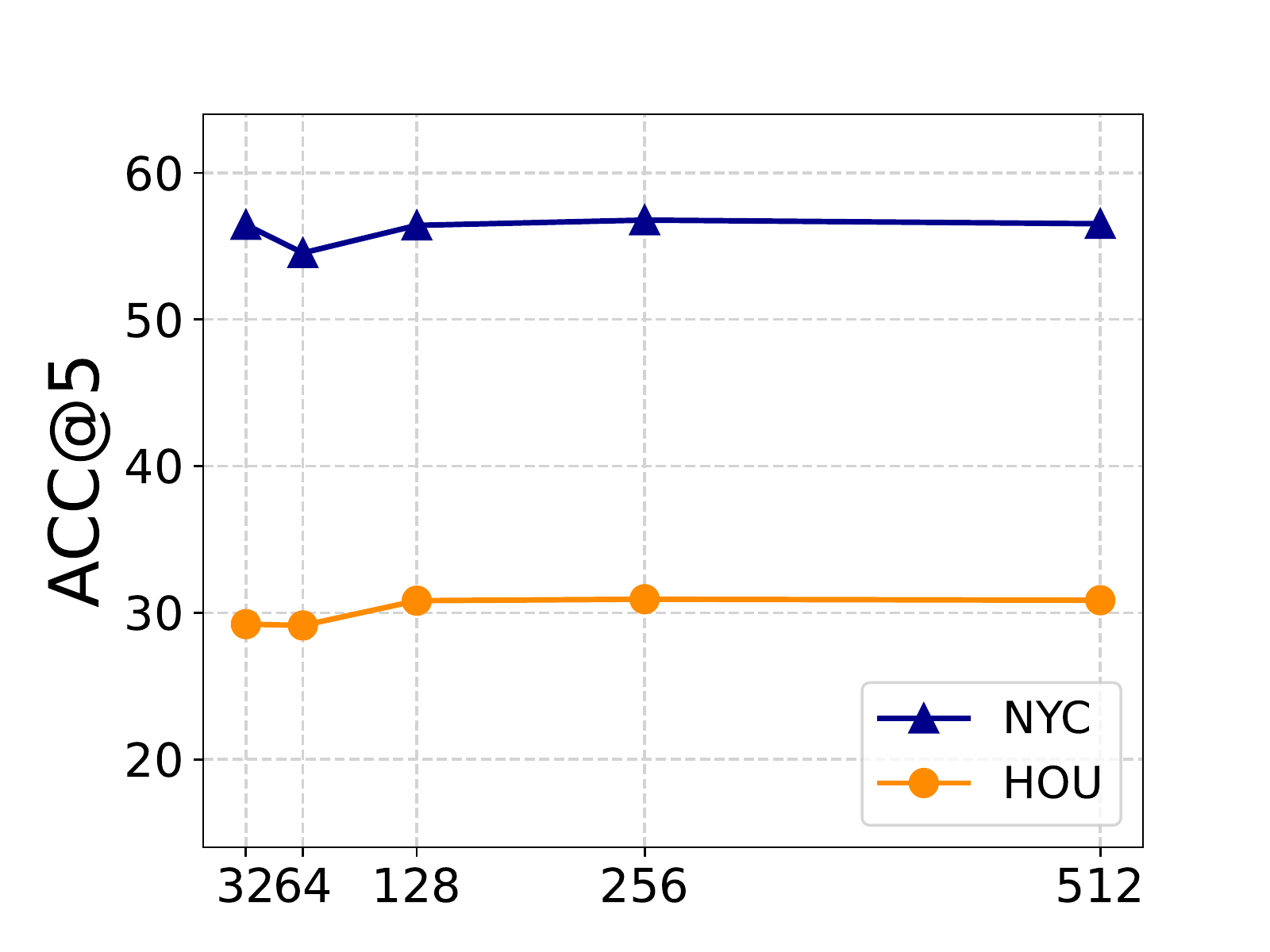}
\hspace{-0.45cm}
\includegraphics[width=0.17\textwidth]
{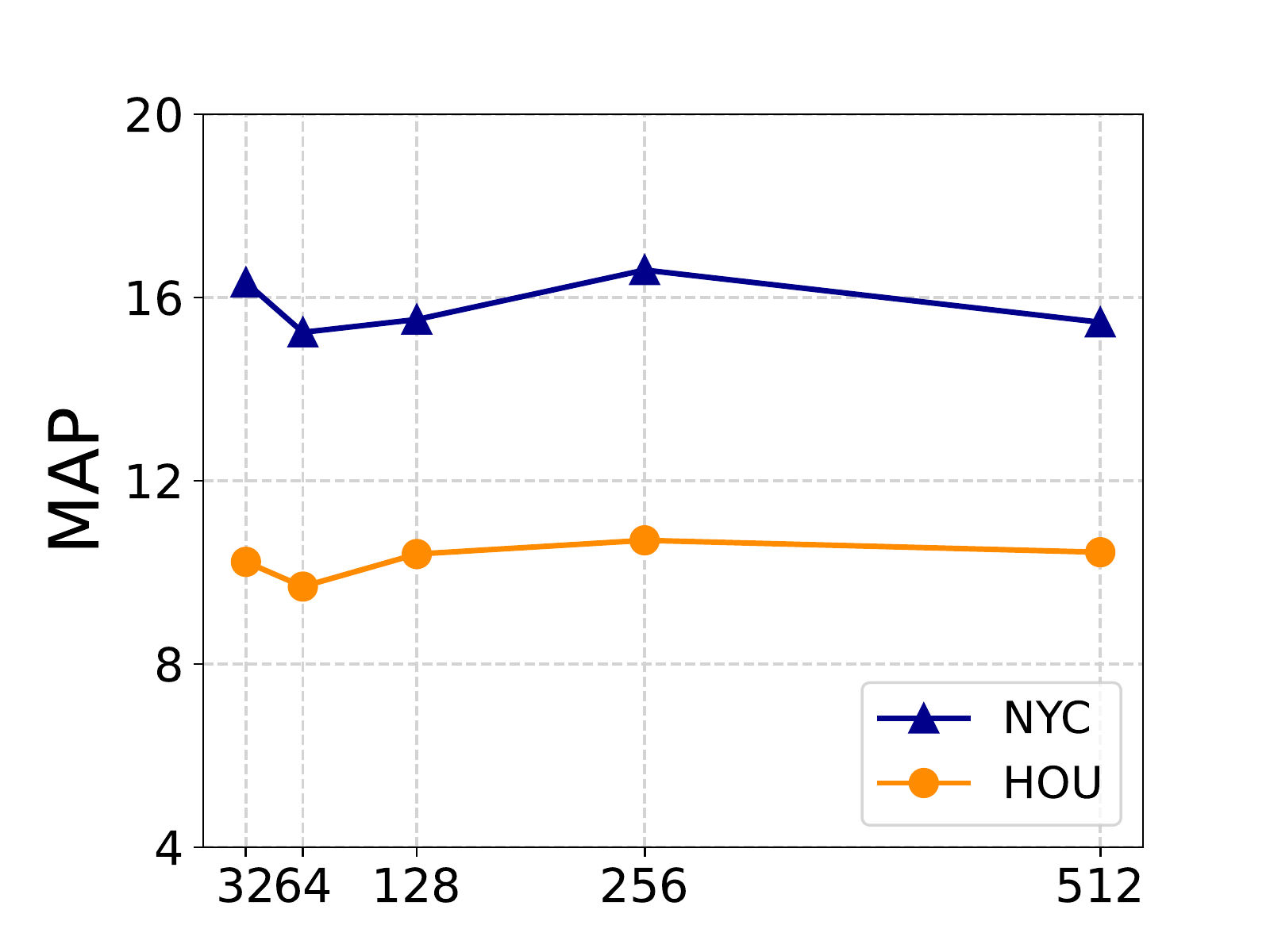}
\caption{The influence of the dimension of $z^s$.}
    \label{fg:static_dim}
\end{figure}

\begin{figure}[ht]
\centering
\includegraphics[width=0.17\textwidth]
{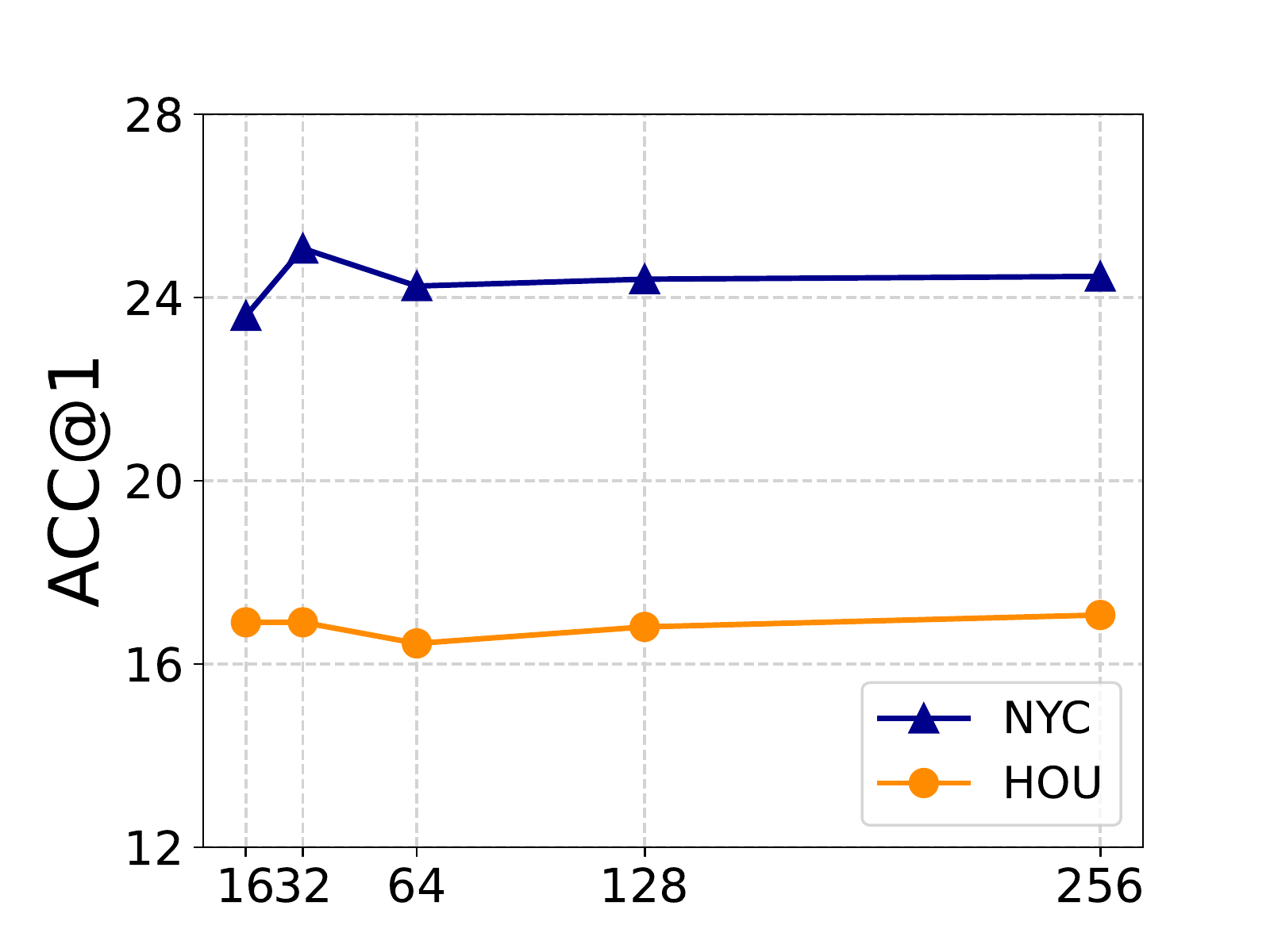}
\hspace{-0.45cm}
\includegraphics[width=0.17\textwidth]
{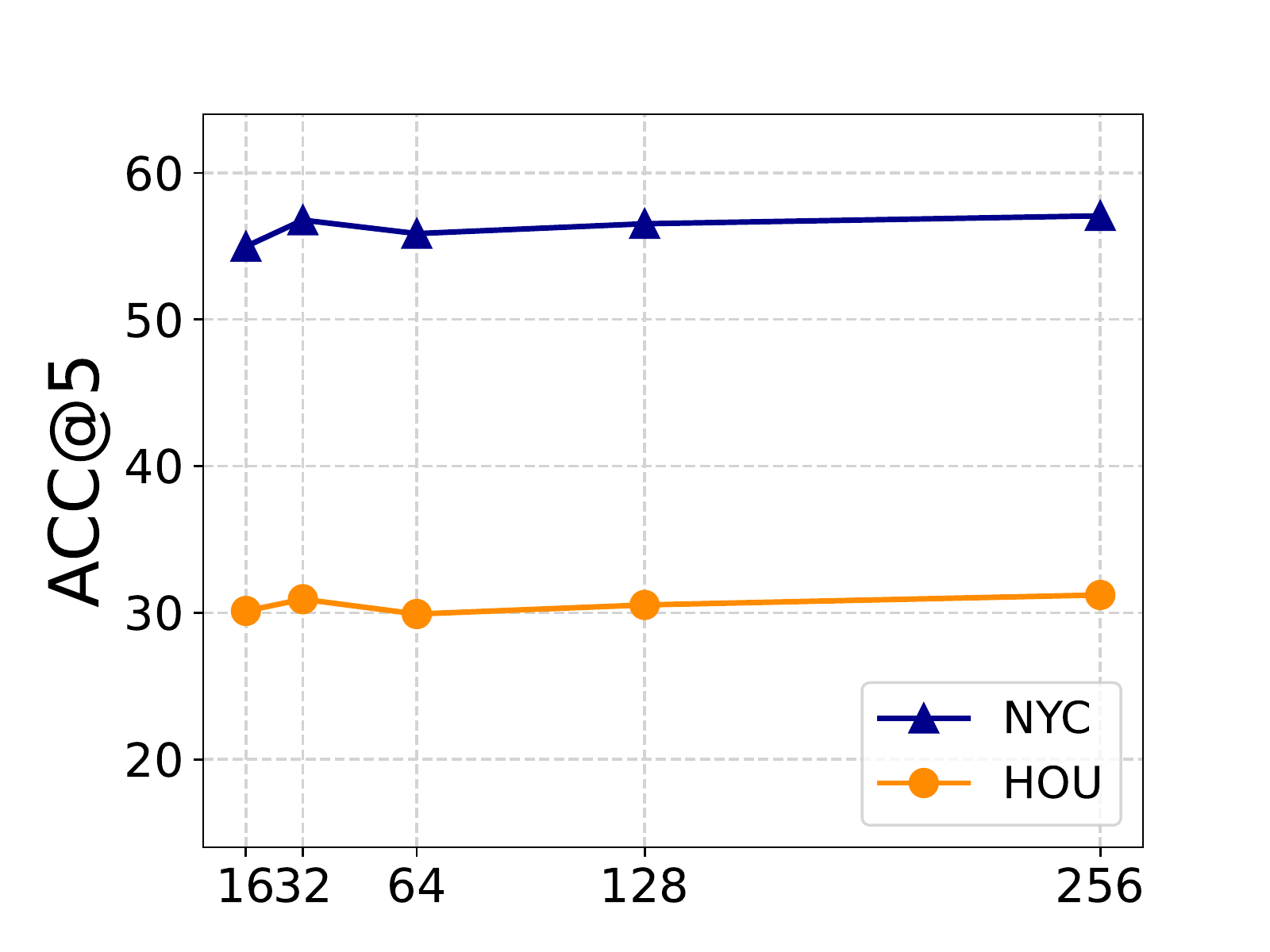}
\hspace{-0.45cm}
\includegraphics[width=0.17\textwidth]
{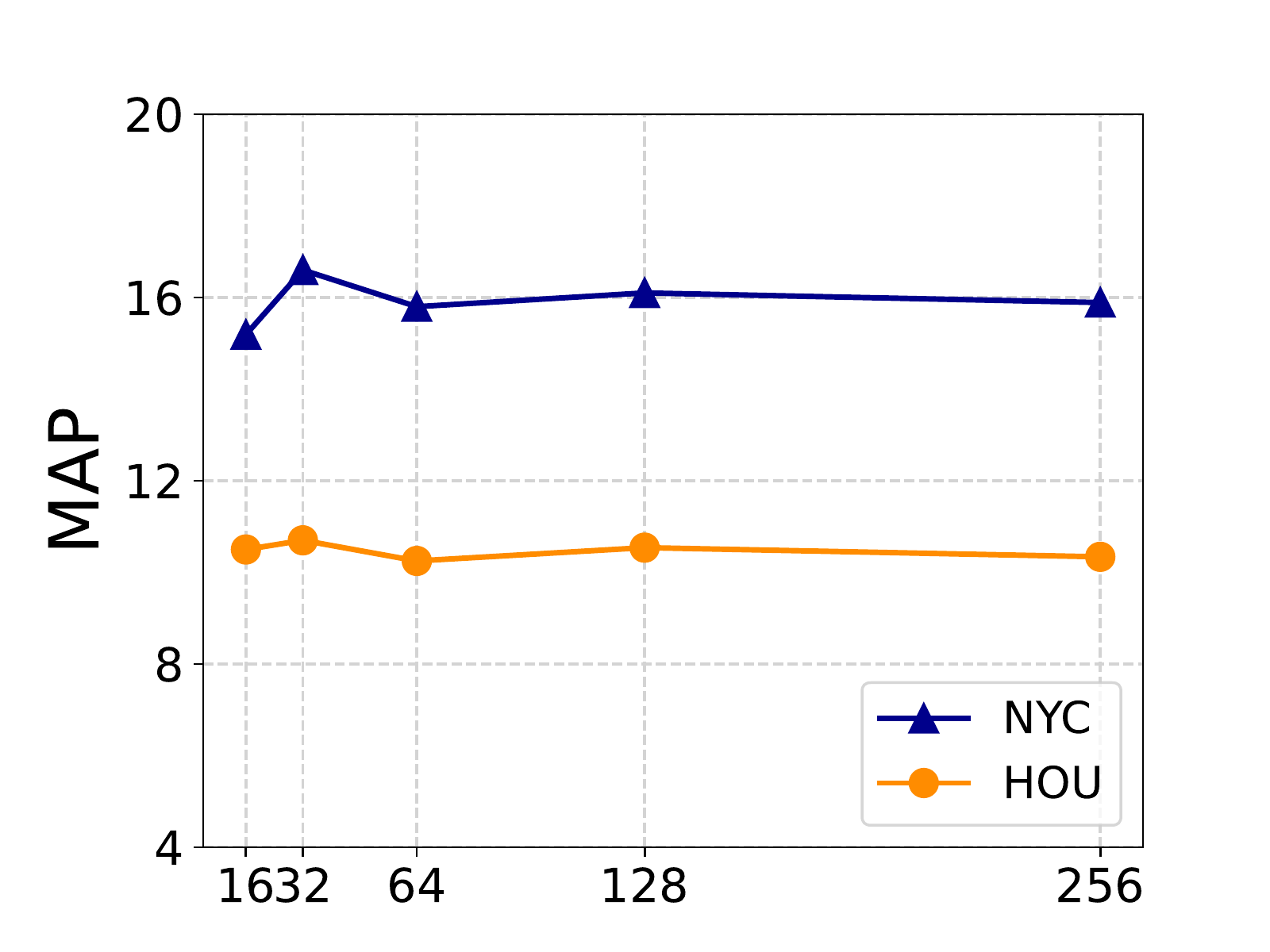}
\caption{The influence of the dimension of $z^r_{1:n}$.}
    \label{fg:dynamic_dim}
\end{figure}

\begin{figure}[ht]
\centering
\includegraphics[width=0.17\textwidth]
{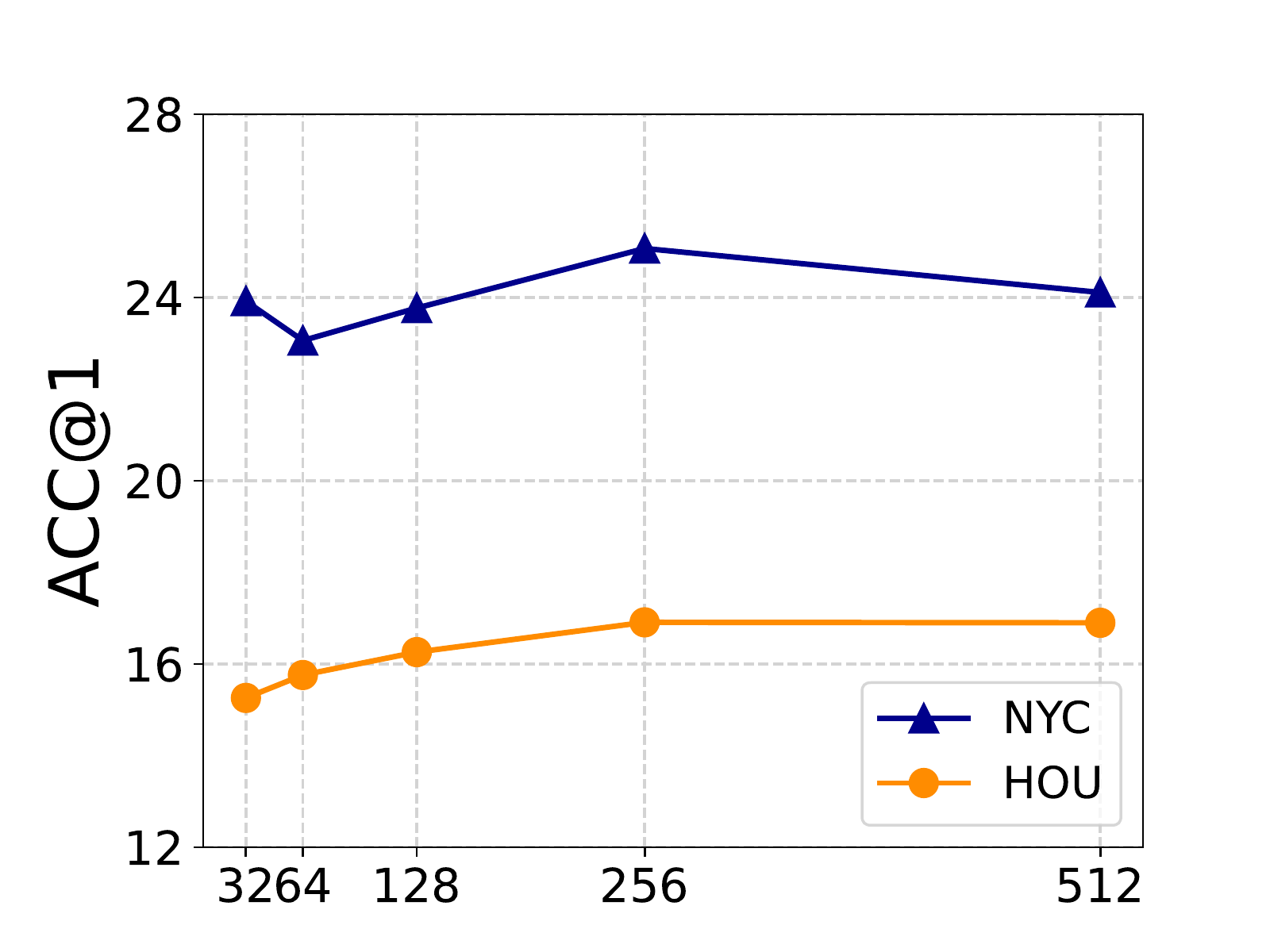}
\hspace{-0.45cm}
\includegraphics[width=0.17\textwidth]
{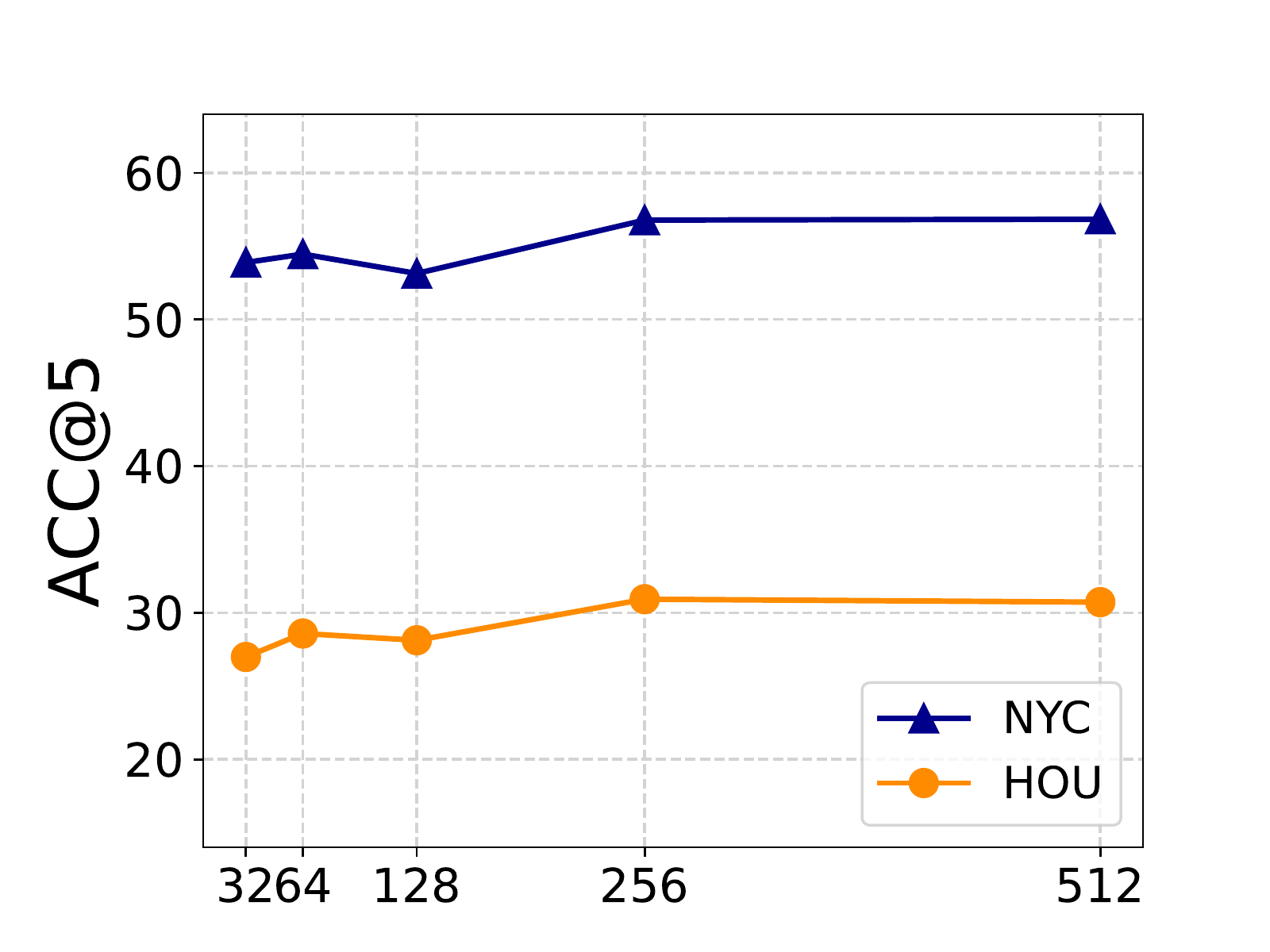}
\hspace{-0.45cm}
\includegraphics[width=0.17\textwidth]
{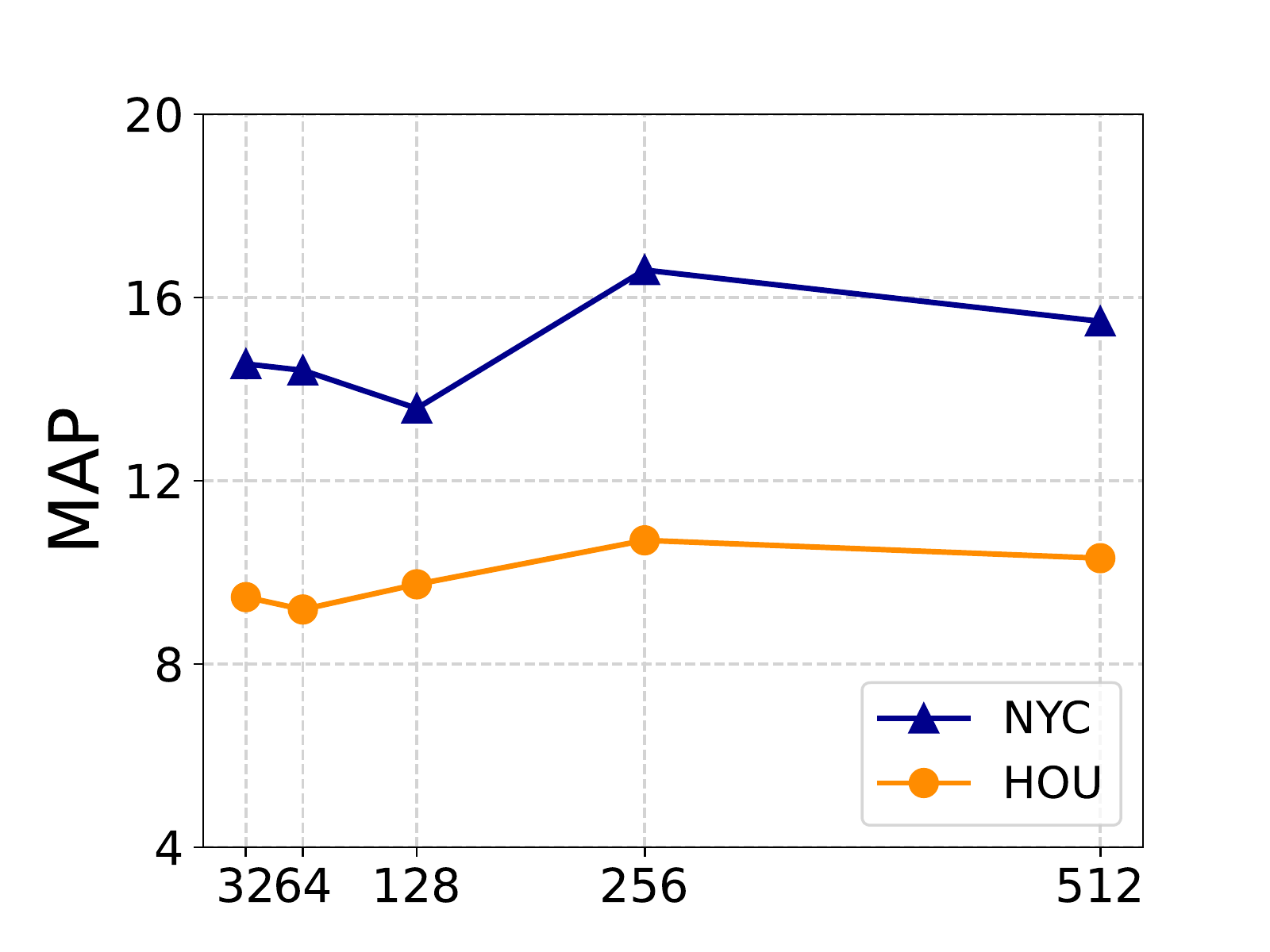}
\caption{The influence of embedding size.}
    \label{fg:emb_size}
\end{figure}

Finally, we investigate the impact of significant hyperparameters in our SSDL to evaluate the model's robustness.
\begin{itemize}[leftmargin=*,align=left]
    \item \textit{Weight coefficients.} %$\alpha$ 
    The objective of our representation learning (cf.~\ref{eq:obj}) contains three coefficients, which would determine the optimization procedure of each relative term. To this end, we generate different combinations of coefficients to investigate their impacts. The results of ACC@1 are shown in Fig.~\ref{fg:weight_coefficients}. We observe that $\gamma$=0.1 obtains better performance than $\gamma$=1 in general. We also find that the larger $\beta$ helps to improve the accuracy of prediction since $\beta$ represents the importance of mutual information between latent variables and trajectories. Finally, the weight coefficient $\alpha$ cannot be too large, otherwise it would constrain the performance.
    \item \textit{Dimension of $z^s$.} Fig.~\ref{fg:static_dim} shows the performance variations of SSDL at different sizes of $z^s$. We find that the larger dimension of $z^s$ does not give us promising results. Hence, for efficiency reasons, we set its dimensionality to 256. 
    \item \textit{Dimension of $z_{1:n}^r$.} Fig.~\ref{fg:dynamic_dim} shows how different dimension of $z_{1:n}^r$ would influence the performance of SSDL. The performance decreases when the dimension of $z_{1:n}^r$ in each time step is larger than 32 and stays stable when the dimension increases. To obtain best performance, we set the dimension of $z_{1:n}^r$ to 32 in our experiments.
    \item \textit{Embedding size.} Embedding size is one of the critical factors affecting task prediction performance. Fig.~\ref{fg:emb_size} presents the effect of the embedding size. We can observe that the performance of SSDL climbs as the embedding size increases, and degrades or stays stable when the embedding size is larger than 256. In our experiments, we set the embedding size to 256.
\end{itemize}

\section{Conclusion}
\label{sec:concl}
In this paper, we present a self-supervised disentanglement learning framework, namely SSDL, to understand human mobility for tackling the next POI prediction problem. In contrast to existing sequential dynamics learning paradigms, SSDL mainly concentrates on disentangling the time-invariant and time-varying factors underlying massive sequential trajectories, which provides us an interpretable perspective to become familiar with human complex mobility patterns. Meanwhile, we present two practical trajectory augmentation strategies to relieve the sparsity issue of check-in data, which also enables the disentanglement of latent representations. Besides, we introduce a flexible graph structure learning method to incorporate multiple heterogeneous collaborative signals from historical check-ins. We believe that several other associated contexts such as social relations and textual data are also easily incorporated into our graph learning. Finally, our extensive experiments on four datasets demonstrate the superiority of SSDL compared to state-of-the-art baselines. As our future work, we plan to investigate the possible more intricate prior assumption during representation learning.

% or


\ifCLASSOPTIONcompsoc
  \section*{Acknowledgments}
\else
  \section*{Acknowledgment}
\fi

This work was supported by the National Natural Science Foundation of China (Grant No.62102326 and No.62072077), the Key Research and Development Project of Sichuan Province under Grant 2022YFG0314, National Science Foundation SWIFT grant 2030249, and Guanghua Talent Project.

\ifCLASSOPTIONcaptionsoff
  \newpage
\fi



%\bibliographystyle{IEEEtran}
\bibliographystyle{IEEEtran}
\bibliography{bib/paper}


%\vfill


\newpage
\section{Appendix}
\label{appendix}

Herein, we provide a theoretical details of our objective. Assume each $z_\tau$ in $z_{1:n}$ is the entangled latent code of check-in $l_\tau$ (i.e., $z_\tau=\{z_{\tau}^r,z^s\}$). We attempt to learn a set of time-varying variables $z_{1:n}^r=\{z_{1}^r,z_{2}^r,\cdots,z_{n}^r\}$ and a time-invariant variable $z^s$ from a given trajectory $l_{1:n}$. According to the Bayes rules and Variational Inference. We have:
\begin{proof}
	\begin{footnotesize}
		\begin{equation}
			\label{eq:proof2}
			\begin{aligned}
				& \log p\left(l_{1: n}\right) \\
				&\geq -KL\left[q\left(z_{1:n}\right) \| p\left(z_{1:n} \mid l_{1: n}\right)\right]+\log p\left(l_{1: n}\right) \\%\mid c_{1:n}
				&=-\mathbb{E}_{q(z_{1:n}\mid l_{1:n})}[ \log \frac{q(z_{1:n})}{p(z_{1:n} \mid l_{1: n})}]+\log p\left(l_{1: n}\right)\\
				&=-\mathbb{E}_{q(z_{1:n}\mid l_{1:n})}[\log q(z_{1:n})-\log p(z_{1:n}\mid l_{1:n})]+\log p(l_{1: n})\\
				&=\mathbb{E}_{q(z_{1:n}\mid l_{1:n})}[\log p(l_{1: n})-\log q(z_{1:n})+\log p(z_{1:n}\mid l_{1:n})]\\
				&=\mathbb{E}_{q(z_{1:n}\mid l_{1:n})}[\log p(l_{1: n})-\log q(z_{1:n})\\
				&+\log \frac{p(c_{1:n} \mid z_{1:n})p(z_{1:n})}{p(c_{1:n})}]\\
				&=\mathbb{E}_{q(z_{1:n}\mid c_{1:n})}[\log p(c_{1: n})-\log q(z_{1:n})\\
				&+\log p(c_{1:n} \mid z_{1:n})+\log p(z_{1:n})-\log p(c_{1:n})]\\
				&=\mathbb{E}_{q(z_{1:n}\mid c_{1:n})}[\log p(c_{1:n} \mid z_{1:n})-(\log q(z_{1:n})-\log p(z_{1:n}))]\\
				&=\mathbb{E}_{q(z_{1:n}\mid c_{1:n})}[\log p(c_{1:n} \mid z_{1:n})]\\
				&-\mathbb{E}_{q(z_{1:n}\mid c_{1:n})}[\log q(z_{1:n})-\log q(z^s)q(z_{1:n}^r)\\
				&+\log q(z^s)q(z_{1:n}^r)-\log p(z_{1:n})]\\
				&=\mathbb{E}_{q(z_{1:n}\mid c_{1:n})}[\log p(c_{1:n} \mid z_{1:n})]\\
				&-\mathbb{E}_{q(z_{1:n}\mid c_{1:n})}[\log q(z_{1:n})-\log q(z^s)q(z_{1:n}^r)]\\
				&-\mathbb{E}_{q(z_{1:n}\mid c_{1:n})}[\log q(z^s)q(z_{1:n}^r)-\log p(z_{1:n})]\nonumber
			\end{aligned}
		\end{equation}
	\end{footnotesize}
	Since $z_{1:n}=(z^s,z^r_{1:n})$, we thus have:
	\begin{footnotesize}
		\begin{equation}
			\label{eq:proof3}
			\begin{aligned}
				& \log p\left(c_{1: n}\right) \\
				&\geq \mathbb{E}_{q(z^s,z^r_{1:n}\mid c_{1:n})}[\log p(c_{1:n} \mid z^s,z^r_{1:n})]\\
				&-\mathbb{E}_{q(z^s,z^r_{1:n}\mid c_{1:n})}[\log q(z^s,z^r_{1:n})-\log q(z^s)q(z_{1:n}^r)]\\
				&-\mathbb{E}_{q(z^s,z^r_{1:n}\mid c_{1:n})}[\log q(z^s)q(z_{1:n}^r)-\log p(z^s,z^r_{1:n})]\\
				&=\mathbb{E}_{q(z^s,z^r_{1:n}\mid c_{1:n})}[\log p(c_{1:n} \mid z^s,z^r_{1:n})]\\
				&-MI_q(z^s;z^r_{1:n})\\
				&-\mathbb{E}_{q(z^s,z^r_{1:n}\mid c_{1:n})}[\log q(z^s)q(z_{1:n}^r)-\log p(z^s,z^r_{1:n})]\\\nonumber
			\end{aligned}
		\end{equation}
	\end{footnotesize}
	Due to the prior assumption $ p(z^s,z^r_{1:n})=p(z^s)p(z^r_{1:n})$, we now have:
	\begin{footnotesize}
		\begin{equation}
			\label{eq:proof4}
			\begin{aligned}
				& \log p\left(c_{1: n}\right) \\
				&\geq\mathbb{E}_{q(z^s,z^r_{1:n}\mid c_{1:n})}[\log p(c_{1:n} \mid z^s,z^r_{1:n})]\\
				&-MI_q(z^s;z^r_{1:n})\\
				&-\mathbb{E}_{q(z^s,z^r_{1:n}\mid c_{1:n})}[\log q(z^s)q(z_{1:n}^r)-\log p(z^s)p(z^r_{1:n})]\\
				&=\mathbb{E}_{q(z^s,z^r_{1:n}\mid c_{1:n})}[\log p(c_{1:n} \mid z^s,z^r_{1:n})]\\
				&-MI_q(z^s;z^r_{1:n})\\
				&-\mathbb{E}_{q(z^s,z^r_{1:n}\mid c_{1:n})}[\log q(z^s)-\log p(z^s)+\log q(z_{1:n}^r)-\log p(z^r_{1:n})]\\
				&=\underbrace{\mathbb{E}_{q(z^s,z^r_{1:n}\mid c_{1:n})}[\log p(c_{1:n} \mid z^s,z^r_{1:n})]}_{1^{st}\ term}\\
				&-\underbrace{MI_q(z^s;z^r_{1:n})}_{6^{st}\ term}\\
				&-\left[\underbrace{\mathbb{E}_{q(z^s,z^r_{1:n}\mid c_{1:n})}[\log\frac{q(z^s)}{p(z^s)}]}_{A}+\underbrace{\mathbb{E}_{q(z^s,z^r_{1:n}\mid c_{1:n})}[\log\frac{q(z_{1:n}^r)}{p(z^r_{1:n})}]}_{B}\right]\\\nonumber
			\end{aligned}
		\end{equation}
	\end{footnotesize}
	
	For part $A$, we have:
	\begin{footnotesize}
		\begin{equation}
			\label{eq:proof5}
			\begin{aligned}
				&A: \mathbb{E}_{q(z^s,z^r_{1:n}\mid c_{1:n})}[\log\frac{q(z^s)}{p(z^s)}]\\
				&=-\mathbb{E}_{q(z^s,z^r_{1:n}\mid c_{1:n})}\left[\log \frac{q(z^s|c_{1:n})}{p(z^s)}-\log \frac{q(z^s|c_{1:n})}{q(z^s)}\right]\\
				&=-\left[\underbrace{KL(q(z^s|c_{1:n})||p(s))}_{2^{st}\ term}-\underbrace{MI_q(z^s,c_{1:n})}_{4^{st}\ term}\right]\nonumber
			\end{aligned}
		\end{equation}
	\end{footnotesize}
	
	For part $B$, we have:
	\begin{footnotesize}
		\begin{equation}
			\label{eq:proof6}
			\begin{aligned}
				&B: \mathbb{E}_{q(z^s,z^r_{1:n}\mid c_{1:n})}[\log\frac{q(z_{1:n}^r)}{p(z^r_{1:n})}]\\
				&=-\mathbb{E}_{q(z^s,z^r_{1:n}\mid c_{1:n})}\left[\log \frac{q(z^r_{1:n}|c_{1:n})}{p(z^r_{1:n})}-\log \frac{q(z^r_{1:n}|c_{1:n})}{q(z^r_{1:n})}\right]\\
				&=-\left[\underbrace{ KL(q(z^r_{1:n}|c_{1:n})||p(z^r_{1:n}))}_{3^{st}\ term}-\underbrace{MI_q(z^r_{1:n},c_{1:n})}_{5^{st}\ term}\right] \nonumber
			\end{aligned}
		\end{equation}
	\end{footnotesize}
	
\end{proof}

\end{document}